\documentclass{article}

\usepackage[preprint]{neurips_data_2024}

\usepackage[utf8]{inputenc} 
\usepackage[T1]{fontenc}    
\usepackage{hyperref}       
\usepackage{url}            
\usepackage{booktabs}       
\usepackage{amsfonts}       
\usepackage{nicefrac}       
\usepackage{microtype}      
\usepackage{xcolor}         

\usepackage{graphicx}%
\usepackage{multirow}%
\usepackage{amsmath,amssymb,amsfonts}%
\usepackage{amsthm}%
\usepackage{mathrsfs}%
\usepackage{textcomp}%
\usepackage{manyfoot}%
\usepackage{booktabs}%
\usepackage{algorithm}%
\usepackage{algorithmicx}%
\usepackage{algpseudocode}%
\usepackage{listings}%
\usepackage{overpic}
\usepackage{subfig}
\usepackage{rotating}
\usepackage{pdflscape}

\title{PlantCamo: Plant Camouflage Detection}

\author{
  Jinyu Yang\textsuperscript{\rm 1}\thanks{Equal contribution.} \quad Qingwei Wang\textsuperscript{\rm 2}\footnotemark[1] \\ 
  \textbf{Feng Zheng}\textsuperscript{\rm 3}\thanks{Corresponding author.} \quad \textbf{Peng Chen}\textsuperscript{\rm 2}\quad \textbf{Ale\v{s} Leonardis}\textsuperscript{\rm 4} \quad 
  \textbf{Deng-Ping Fan}\textsuperscript{\rm 5}
  \\
  \textsuperscript{\rm 1} Tapall.ai \quad
  \textsuperscript{\rm 2} China Three Gorges University \quad
  \textsuperscript{\rm 3} Southern University of Science and Technology \\
  \textsuperscript{\rm 4} University of Birmingham \quad
  \textsuperscript{\rm 5} Nankai University \\
}

\begin{document}

\maketitle

\begin{abstract}
Camouflaged Object Detection (COD) aims to detect objects with camouflaged properties.
Although previous studies have focused on natural (animals and insects) and unnatural (artistic and synthetic) camouflage detection, plant camouflage has been neglected. 
However, plant camouflage plays a vital role in natural camouflage. Therefore, this paper introduces a new challenging problem of Plant Camouflage Detection (PCD). To address this problem, we introduce the \textit{PlantCamo} dataset, which comprises 1,250 images with camouflaged plants representing 58 object categories in various natural scenes. To investigate the current status of plant camouflage detection, we conduct a large-scale benchmark study using 20+ cutting-edge COD models on the proposed dataset. 
Due to the unique characteristics of plant camouflage, including holes and irregular borders, we developed a new framework, named PCNet, dedicated to PCD.
Our PCNet surpasses performance thanks to its multi-scale global feature enhancement and refinement. Finally, we discuss the potential applications and insights, hoping this work fills the gap in fine-grained COD research and facilitates further intelligent ecology research. 
All resources will be available on \url{https://github.com/yjybuaa/PlantCamo}.
\end{abstract}

\section{Introduction}\label{intro}



Camouflaged Object Detection (COD) has sparked a flurry of interest in the field of computer vision, spurred by a wave of pioneering studies~\cite{lamdouar2023making,penet,UQFormer,FEDER,EVP,CRNet,MFFN}. In the area of computer vision, COD aims to distinguish between two distinct types of camouflaged objects in real-world environments: naturally and unnaturally camouflaged objects~\cite{CAMO,COD10K,NC4K}. Naturally camouflaged objects refer to animals or insects that cleverly hide themselves from their predators, while unnaturally camouflaged ones pertain to human-made objects like objects cleverly disguised by synthetic texture patterns such as body painting or camouflage clothing.
In comparison to animals and humans, studies of camouflage in plants have often been overlooked.
However, biologists have discovered and verified that a multitude of plants utilize camouflage to protect themselves from enemies, such as herbivores and other antagonists~\cite{plant}.
Remarkably, plants deploy different strategies to camouflage themselves, similar to animals. The plant camouflage strategies~\cite{plant} mainly include:
1) \textbf{Background matching} - plants blended with the colors and shapes of their natural habitat.
2) \textbf{Disruptive coloration} - markings that create the illusion of false edges and boundaries, making it harder for the observer to perceive the plant's true outline.
3) \textbf{Masquerade} - plants cleverly resemble something else, often an object that a predator might overlook, such as a stone or twig. Notable examples include lithops, cacti, passion vines, and mistletoes.
4) \textbf{Decoration} - plants accumulate material from their environment to cover themselves. For instance, some coastal and dune plants become covered with sand, making them less conspicuous.
Furthermore, the detection of camouflaged plants has been demonstrated to effectively contribute to the preservation of endangered species and the discovery of new species~\cite{commercial}. Nevertheless, plant camouflage remains under-explored in the field of computer vision, despite its potential as a crucial component in comprehending concealed scenes.

\begin{figure}[t]
    \begin{center}
        \begin{overpic}[width=.95\linewidth]{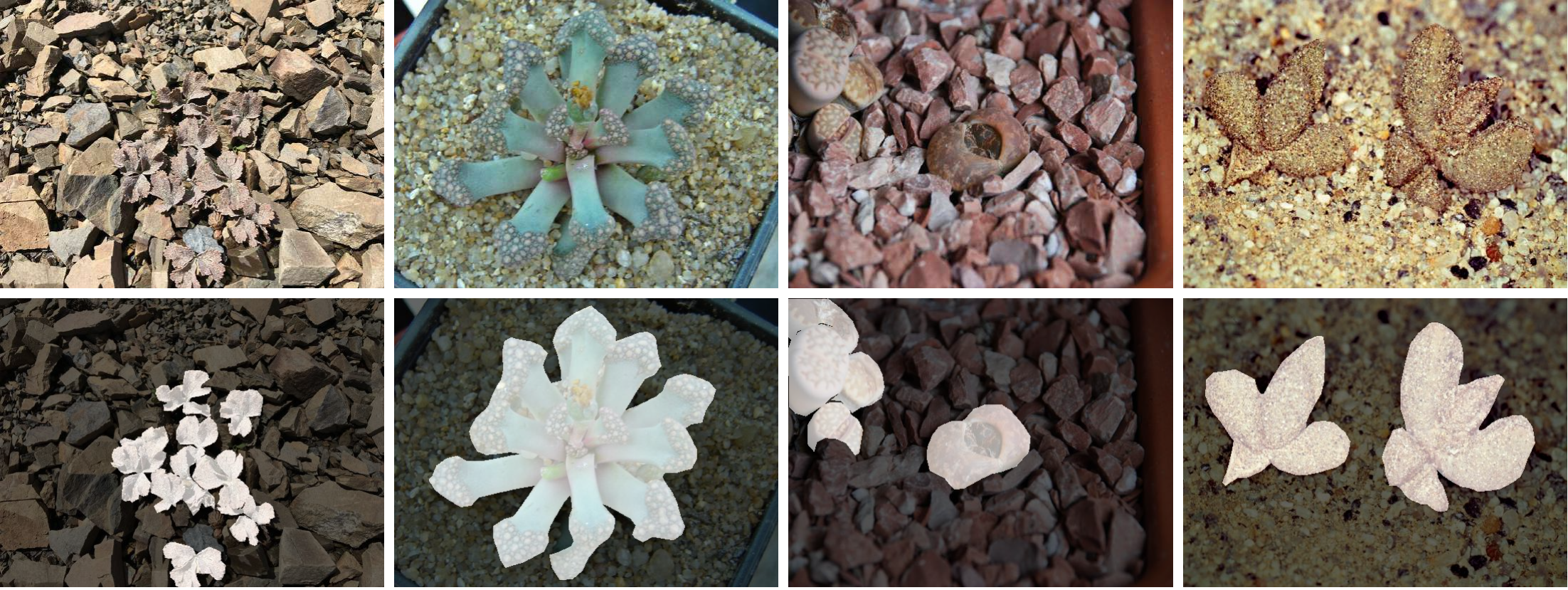}
    \put(1.5,38.5){\scriptsize \textbf{(a)}}
    \put(4.5,38.5){\scriptsize \textbf{Background Matching}}
    \put(26.5,38.5){\scriptsize \textbf{(b)}}
    \put(29.5,38.5){\scriptsize \textbf{Disruptive Coloration}}
    \put(54.5,38.5){\scriptsize \textbf{(c)}}
    \put(57.5,38.5){\scriptsize \textbf{Masquerade} }
    \put(80.5,38.5){\scriptsize \textbf{(d)}}
    \put(83.5,38.5){\scriptsize \textbf{Decoration}} 
    \put(-2.5,26){\scriptsize \rotatebox{90}{\textbf{Image}}} 
    \put(-2.5,7){\scriptsize \rotatebox{90}{\textbf{Mask}}} 
    \end{overpic}        
    \end{center}
    \captionof{figure}{
Visualized examples and corresponding mask annotations in \textbf{PlantCamo dataset}. Our dataset covers different kinds of plant camouflage: (a) background matching; (b) disruptive coloration; (c) masquerade; and (d) decoration.
    }
    \label{dataoverview}
\end{figure}
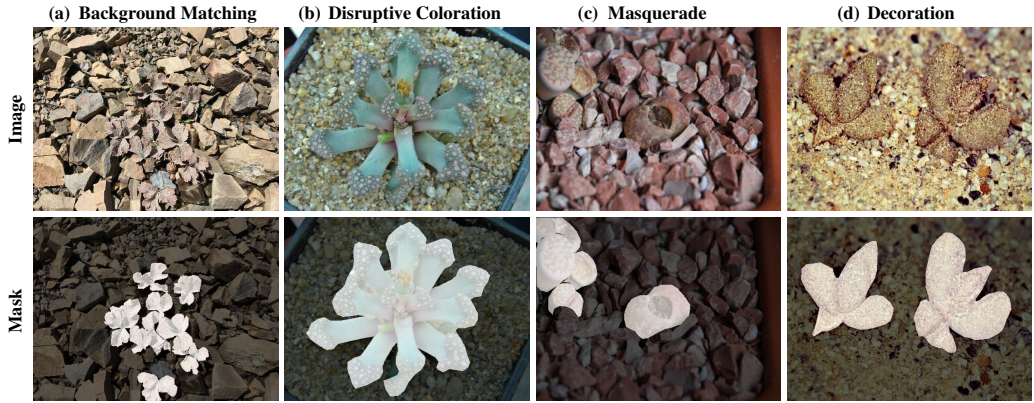

We observe that the presentation of plant camouflage diverges from animal or synthetic camouflage, despite employing similar camouflage strategies. 
For example, plants utilizing camouflage are predominantly succulents, like lithops, which mimic the appearance of pebbles to avoid detection and subsequent feeding by insects~\cite{dyer2021plant}.
Furthermore, these camouflaged plants thrive in environments where resources are limited, such as barren plateau areas. 
In these scenarios, where nutrients are sparse and animals actively search for food, only very few plants can survive. 
Thus, plants in these environments have evolved to develop intricate color patterns that further enhance their camouflage and help them evade detection by animals~\cite{dyer2021plant}. 
This plant camouflage phenomenon provokes a series of new questions for the field of COD research:
1) What are the visual expressions of plant camouflage?
2) Are current COD models effective for plant camouflage?
3) How will plant camouflage datasets impact data-driven COD models?

In addressing these inquiries, we present the first dataset and benchmark for Plant Camouflage Detection (PCD), intending to identify plants based on their visual camouflage within their background.
To begin, we compile the PlantCamo dataset, consisting of 1,250 images showcasing 58 kinds of camouflaged plants. Within PlantCamo, all four types of plant camouflage are encompassed, as shown in Fig.~\ref{dataoverview}. 
Moreover, we furnish pixel-, bounding box-, attribute-, and category-level annotations, facilitating a thorough comprehension of plant camouflage.

To study the capabilities of current state-of-the-art (SoTA) COD models, we conduct a comprehensive benchmark study based on the PlantCamo dataset, using more than 20 cutting-edge COD models.
Interestingly, we find current advanced COD models perform worse on PlantCamo than previous benchmarks, due to the different intrinsic characteristics in plant camouflage.
Building upon the distinctive features of camouflaged plants, we introduce a specialized framework, PCNet, for plant camouflage detection. 
On the one hand, PCNet acquires global information through bottom-up fusion to accurately detect and locate camouflaged plants. 
On the other hand, features are refined through top-down fusion for precise boundary segmentation. 
Also, feature optimization for precise prediction is accomplished through a feedback strategy.
Experiments demonstrate that the proposed PCNet achieves advanced performance on the PlantCamo, compared with SoTAs, demonstrating its ability on plant camouflage detection.
The overall contribution of this paper is four-fold:

\begin{enumerate}
\item \textbf{New exploration.} Different from existing works on COD, we newly explore plant camouflage detection, which is an important part of natural camouflage and bridges the gap between current COD research and real-world applications.
\item \textbf{New dataset.} We collect a new image dataset dedicated to plant camouflage detection, namely \textit{PlantCamo}.
It contains 1,250 images covering 58 categories of camouflaged plants.
Each image is hierarchically annotated to enable multiple vision research tasks.
\item \textbf{New benchmark.} We give a thorough benchmark study for camouflaged plant detection on the proposed PlantCamo dataset. Based on the investigation, we give in-depth analysis and discussion, to inspire new ideas for PCD.
\item \textbf{New method.} We propose a baseline dedicated to plant camouflage detection, namely \textit{PCNet}.
Through the novel Multi-scale Global Feature Enhancement (MGFE) and Multi-scale Feature Refinement (MFR) modules, inspired by the intrinsic characteristic of plant camouflage, PCNet outperforms the SoTAs by a significant margin.
\end{enumerate}

\begin{table*}[t]
\setlength\tabcolsep{5pt}
\renewcommand{\arraystretch}{1.0}
\small
\centering
\setlength{\belowcaptionskip}{0cm} 
 \caption{Dataset comparison.}
\centering
\begin{tabular}{l|l|l|l|l|ll} 
\toprule
Dataset & Chameleon~\cite{chameleon} &CAMO~\cite{CAMO} &COD10K~\cite{COD10K} &NC4K~\cite{NC4K} &\textbf{PlantCamo}\\
\midrule
Venue & - &CVIU &CVPR &CVPR &\textbf{-}\\
Year & 2018 &2019  &2020 & 2021 & \textbf{2024}\\
Scope &Animal & Animal\&unnatural& Animal\&unnatural& Animal\&unnatural &\textbf{Plant}\\
\#Image &76 &1,250 &10,000 &4,121 &\textbf{1,250}\\
\#Class &- &8 &78 &- &\textbf{58}\\
\#Attr. &- &7 &7 &- &\textbf{10}\\
\bottomrule
\end{tabular}
\label{tab:dataset}
\end{table*}

\begin{figure*}[t]
	\centering
    \small
	\begin{overpic}[width=0.99\linewidth]{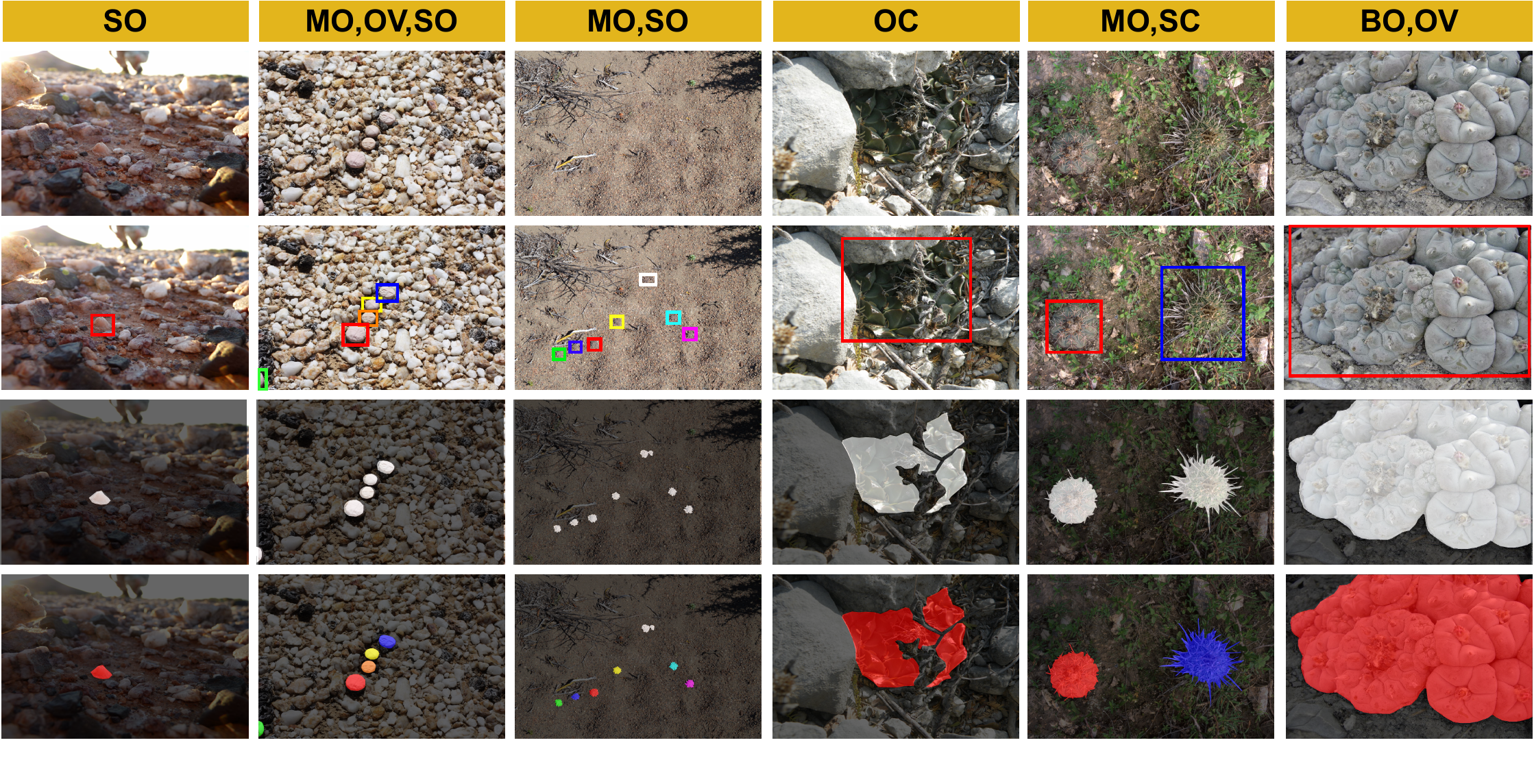} 
    \put(-5,37.5){\scriptsize \rotatebox{90}{\textbf{Image-level}}}
    \put(-2.5,38){\scriptsize \rotatebox{90}{\textbf{Attributes}}}

    \put(-5,26.5){\scriptsize \rotatebox{90}{\textbf{Bounding}}}
    \put(-2.5,29.5){\scriptsize \rotatebox{90}{\textbf{Box}}}
    \put(-5,16.5){\scriptsize \rotatebox{90}{\textbf{Object}}}
    \put(-2.5,14.5){\scriptsize \rotatebox{90}{\textbf{Annotation}}}

    \put(-5,5){\scriptsize \rotatebox{90}{\textbf{Instance}}}
    \put(-2.5,3.5){\scriptsize \rotatebox{90}{\textbf{Annotation}}}

     \put(6.5,1){\small (A)}
    \put(23.5,1){\small (B)}
    \put(40,1){\small (C)}
    \put(56.5,1){\small (D)}
    \put(73.5,1){\small (E)}
    \put(90.5,1){\small (F)}
    \end{overpic}    
	\caption{Annotated examples in the proposed PlantCamo. For each image, we offer different annotations, which include image-level attributes (1st row), bounding boxes (2nd row), object annotation (3rd row), and instance annotation (4th row). Zoom in for details.}
 \label{annotation}
\end{figure*}
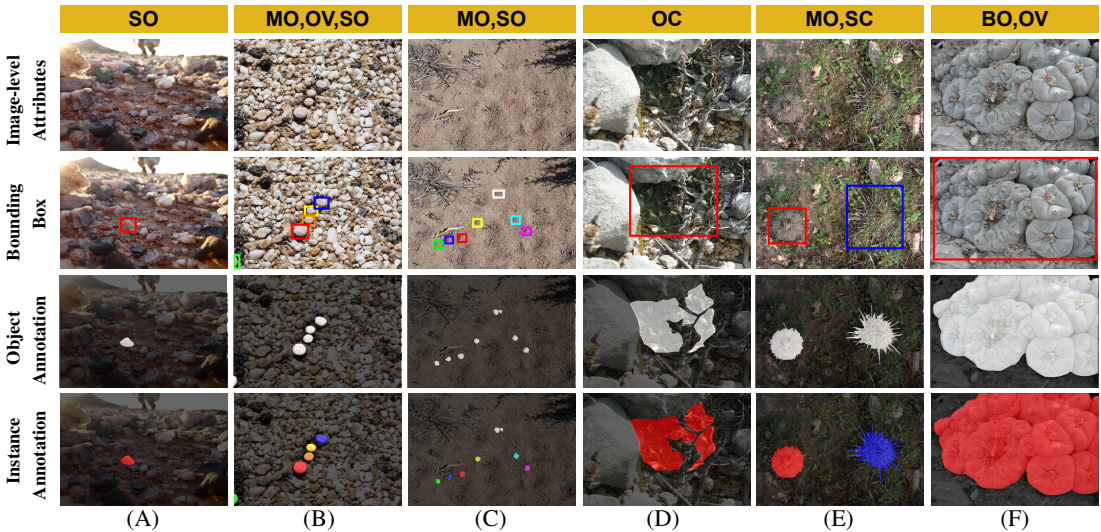

\begin{table*}[t]
\renewcommand{\arraystretch}{1.0}
\small
\centering
\setlength{\belowcaptionskip}{0cm}  
\caption{Data attribute and corresponding description.}
\label{attribute}
\begin{tabular}{c|l|l}
\toprule
Abb. & Attribute & Description\\
\midrule
\textbf{BM} & \textit{Background matching} & The target blends with the colors of surroundings~\cite{plant}. \\
\midrule
\textbf{DC} & \textit{Disruptive coloration} &Appearance of false edges and boundaries~\cite{plant}.\\
\midrule
\textbf{MQ} & \textit{Masquerade} & Targets looks like something else, \textit{e.g.}, stone or twig~\cite{plant}.\\
\midrule
\textbf{DR} & \textit{Decoration} & Targets are covered by materials from the environment~\cite{plant}.\\
\midrule
\textbf{MO} & \textit{Multiple objects} &Number of objects in an image is larger than 1~\cite{COD10K}.\\
\midrule
\textbf{SC} & \textit{Shape complexity} &
The object has complex boundaries, \textit{e.g.}, irregular or jagged~\cite{COD10K}.\\
\midrule
\textbf{OC} & \textit{Occlusion} &Object is partially occluded~\cite{COD10K}.\\
\midrule
\textbf{BO} & \textit{Big object} &Ratio between object area and image area $\geq$ 0.5~\cite{COD10K}. \\
\midrule
\textbf{SO} & \textit{Small object} &Ratio between object area and image area $\leq$ 0.1~\cite{COD10K}.\\
\midrule
\textbf{OV} & \textit{Out-of-view} &Object is clipped by image boundaries~\cite{COD10K}.\\
\bottomrule
\end{tabular}
\end{table*}

\section{Related Work}\label{sec:relatedwork}

\subsection{COD in Biology \& Evolution}
In the field of biology, camouflage has long played a crucial role as a defense strategy among animals and has been a subject of extensive study with evolution for over 100 years~\cite{plant,Color_of_Animal_plants}. 
However, compared to the extensive knowledge of how animals camouflage themselves, research on plant camouflage remains relatively limited. This is primarily due to the fact that plants require chlorophyll to survive through photosynthesis, which typically gives them a green hue. 
However, in recent years, plant camouflage has begun to attract a growing amount of attention owing to related research and evidence~\cite{dyer2021plant}. 
Biologists have observed that certain plants have adapted their colors, shapes, and patterns to match their environments, making them difficult to detect and predate upon~\cite{niu2017divergence,Niu2014AlpineSP}. Plant camouflage holds significant importance in both biology and evolution, particularly in terms of visual systems and survival strategies. Unfortunately, this important topic has remained largely unexplored in computer vision research, especially in the field of COD.

\subsection{COD in Computer Vision}
COD is proposed as a class-agnostic task to distinguish the objects with camouflage characteristics from the background~\cite{COD10K}.
The high appearance similarities between the target object and the background make COD far more challenging than generic object detection.

\textbf{Datasets.}
In recent years, multiple datasets have been collected for COD tasks.
The first COD dataset is Chameleon~\cite{chameleon}, which contains only 76 camouflaged animal images with object-level annotations.
It is collected via a Google search using ``concealed animal'' as a keyword.
After that, CAMO~\cite{CAMO} dataset is proposed, which involves both natural and unnatural camouflage.
It contains 1,250 images with a split of 1,000 for training and 240 for testing.
Then, Fan \textit{et al.}~\cite{COD10K} released the COD10K dataset, which is the largest public dataset for COD tasks.
COD10K contains 10,000 images, including both generic and camouflaged images, of which 60\% is for training and 40\% is for testing.
The largest test set for COD up to now is NC4K~\cite{NC4K}, including $\sim$4,000 camouflaged images for evaluating COD models.
However, existing datasets mainly involve animal and synthetic camouflage, while plant camouflage is almost neglected and lacks specialized datasets.
In Tab.~\ref{tab:dataset}, we compare the collected PlantCamo with existing COD datasets.
As shown, PlantCamo is the first dataset focusing on plant camouflage detection, which fills the gap of fine-grained natural camouflage detection.

\textbf{Methods.}
To recognize camouflaged objects, it has been numerous efforts in the area of computer vision~\cite{JSCOD,TINet}.
The pioneering work ANet~\cite{CAMO} detects camouflaged objects by utilizing the awareness of camouflage as prior.
Then, various related cues are involved as assistance to COD, \textit{e.g.}, frequency~\cite{FDNet,FRINet,FPNet}, depth~\cite{depthcod,popnet,dacod}, and boundary~\cite{boundary1,BSANet,findnet}.
Recent works tried to constrain the attention on the camouflaged objects, with iterative refinement~\cite{SegMaR}, context-aware cross-level fusion~\cite{C2FNet}, mixed-scale integration~\cite{ZoomNet}, bio-inspired mechanism~\cite{iclr24}, and masked separable attention~\cite{camoformer}.
Compared with animal camouflage, plant camouflage shows different expressions as they do not have the visual system and movement ability.
Also, whether the advanced COD methods are effective on plants is uninvestigated.

\begin{figure*}[t]
  \centering
\begin{overpic}[width=\textwidth]{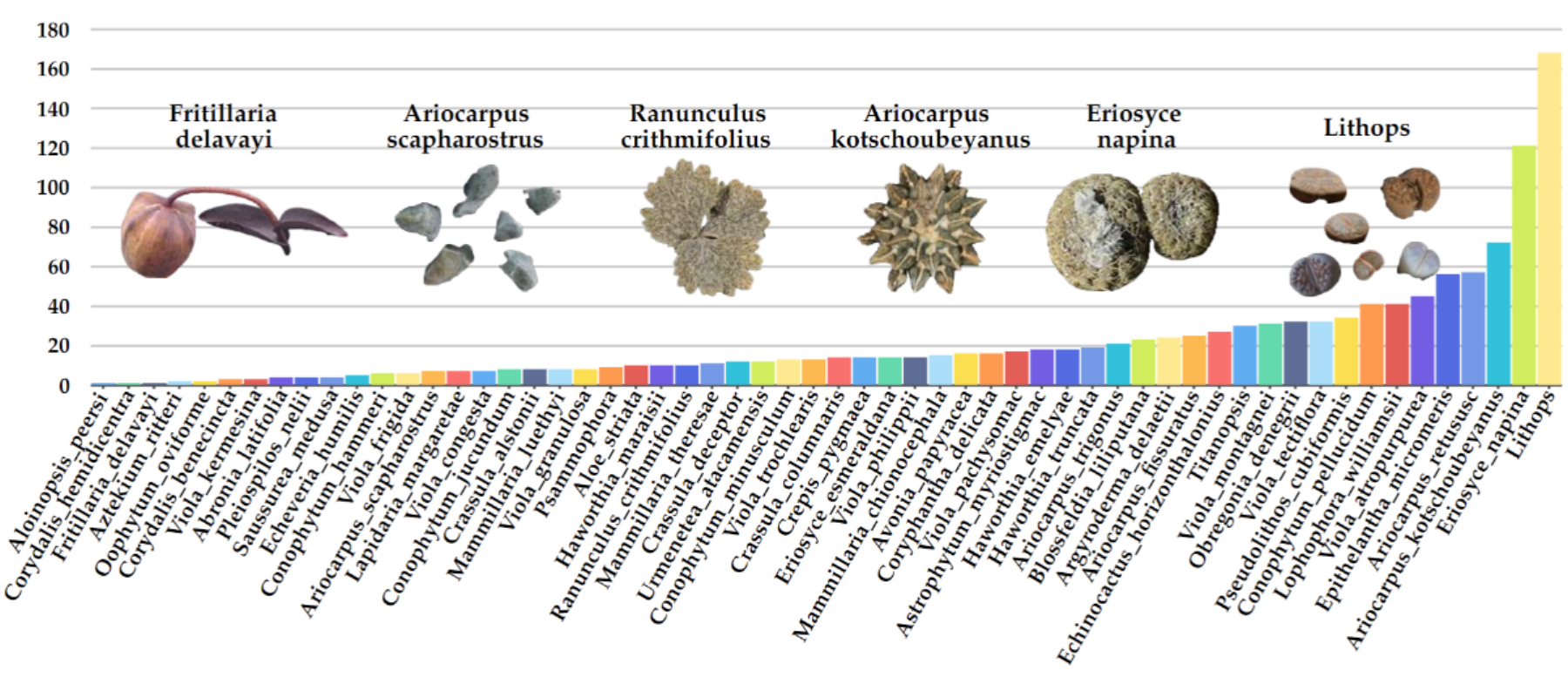}
  \end{overpic}
  \caption{Histogram distribution for the plant categories in PlantCamo. We visualize some representative camouflaged plants.}
    \label{fig:category}
\end{figure*}

\begin{figure*}[t]
  \centering
    \subfloat[Camouflage strategy distribution.]
  {
  \begin{overpic}[width=0.32\textwidth]{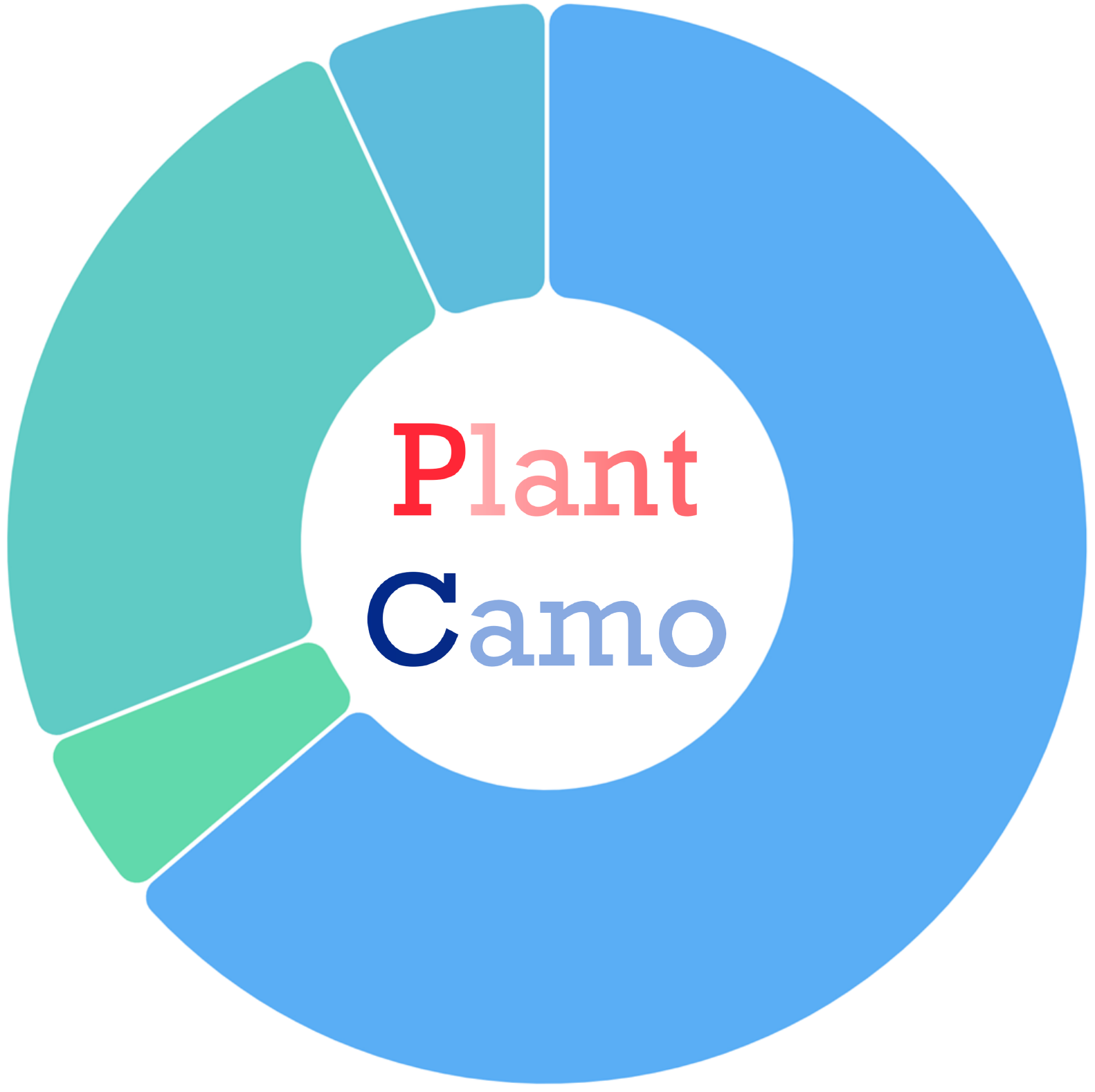}
\put(77,44){\footnotesize {\textbf{BM}}}
\put(72,37){\footnotesize {\textbf{63.79\%}}}
\put(38,90){\footnotesize {\textbf{DC}}}
\put(36,83){\footnotesize {\textbf{6.90\%}}}
\put(13,69){\footnotesize {\textbf{MQ}}}
\put(9,62){\footnotesize {\textbf{24.14\%}}}
\put(15,35){\footnotesize {\textbf{DR}}}
\put(12,28.5){\footnotesize {\textbf{5.17\%}}}
  \end{overpic} 
  \label{fig:subfig11}} 
    ~
\subfloat[Visual attribute distribution.]
  {
  \begin{overpic}[width=0.32\textwidth]{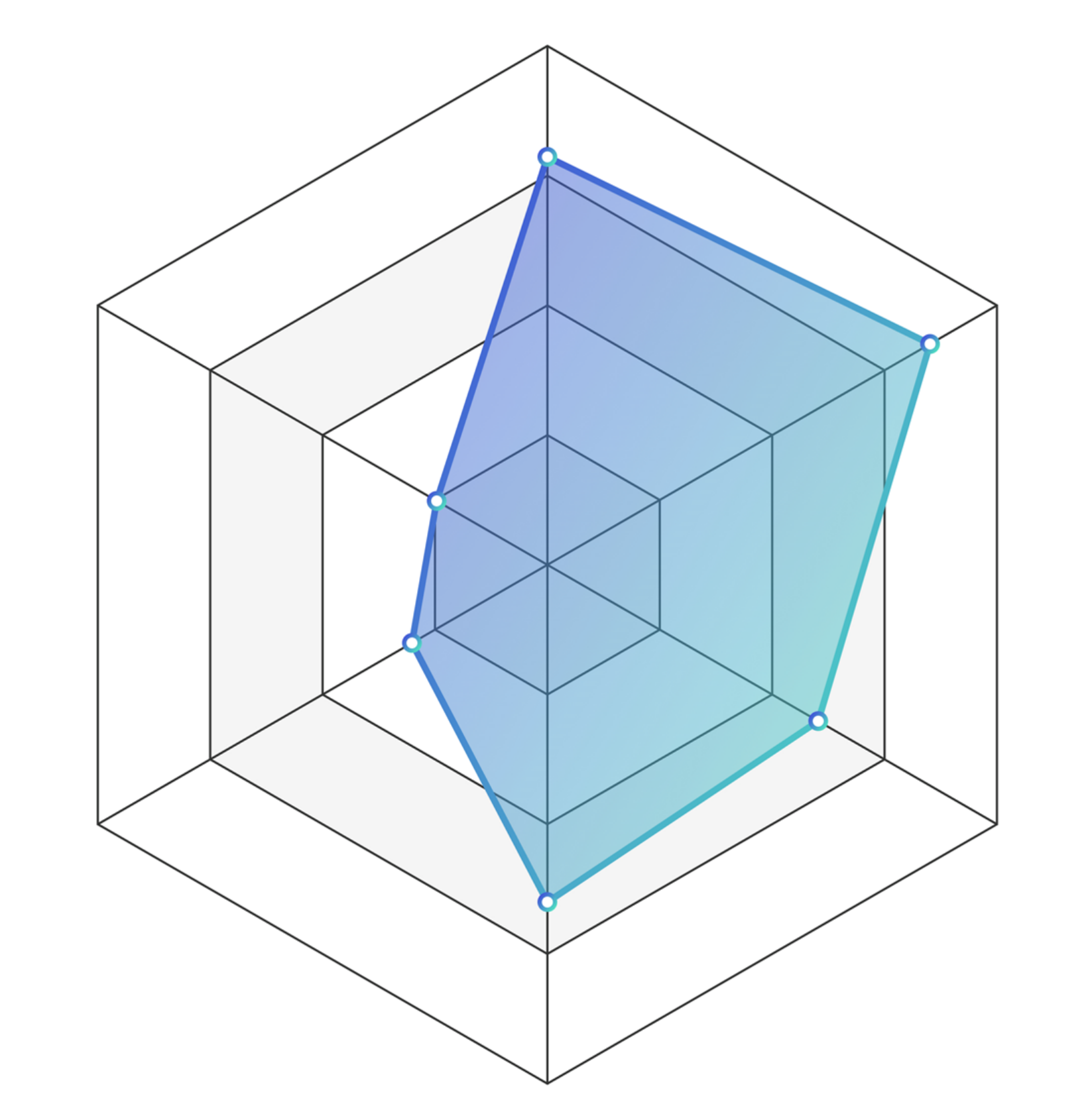}
\put(45,-1.5){\footnotesize {\textbf{BO}}}
\put(44,97){\footnotesize {\textbf{MO}}}
\put(-0.5,22.5){\footnotesize {\textbf{SO}}}
\put(-0.5,73){\footnotesize {\textbf{OV}}}
\put(90,73){\footnotesize {\textbf{SC}}}
\put(90,22.5){\footnotesize {\textbf{OC}}}

\put(45,14){\scriptsize {\textbf{520}}}
\put(44,87){\scriptsize {\textbf{629}}}
\put(27,37){\scriptsize {\textbf{241}}}
\put(28,55){\scriptsize {\textbf{197}}}
\put(80,70){\scriptsize {\textbf{681}}}
\put(72,29){\scriptsize {\textbf{482}}}
  \end{overpic}
  \label{fig:subfig7}}
  ~
  \subfloat[Multi-dependencies in attributes.]
  {
  \begin{overpic}[width=0.32\textwidth]{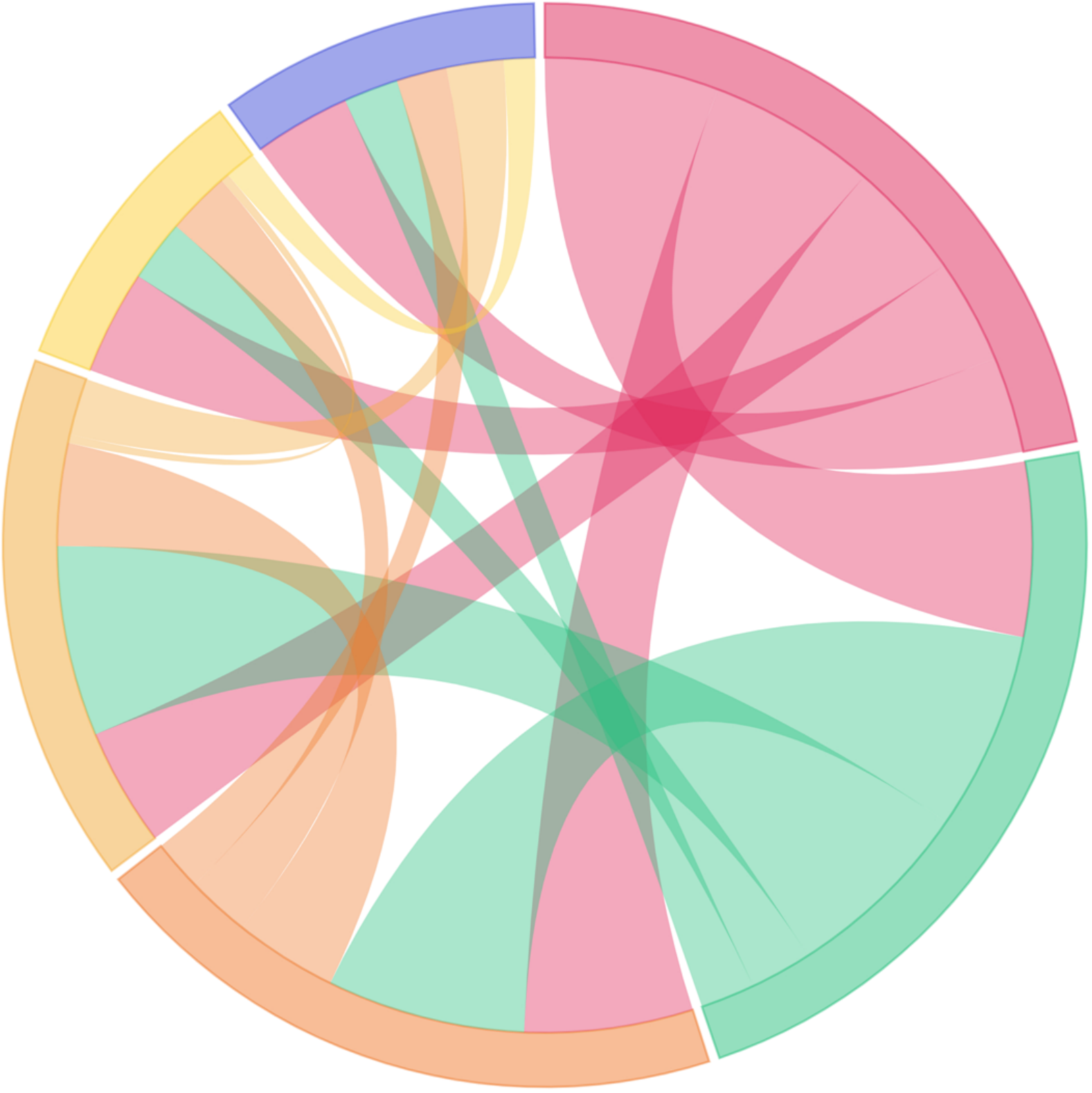}

  \put(-5,45){\footnotesize \rotatebox{-80}{\textbf{BO}}}
  \put(28,0){\footnotesize \rotatebox{-15}{\textbf{OC}}}
  \put(90,18){\footnotesize \rotatebox{60}{\textbf{SC}}}
  \put(78,92){\footnotesize \rotatebox{-40}{\textbf{MO}}}
  \put(28,96){\footnotesize \rotatebox{20}{\textbf{OV}}}
  \put(5,78){\footnotesize \rotatebox{53}{\textbf{SO}}}
   \end{overpic}
  \label{fig:subfig8}} 
  \caption{Dataset statistics of the proposed PlantCamo. BM = background matching, DC = disruptive coloration, MQ = masquerade, DR = decoration, MO = multiple objects, SC = shape complexity, OC = occlusion, BO = big object, SO = small object, and OV = out-of-view.}
  \label{fig:subfig9}
\end{figure*}
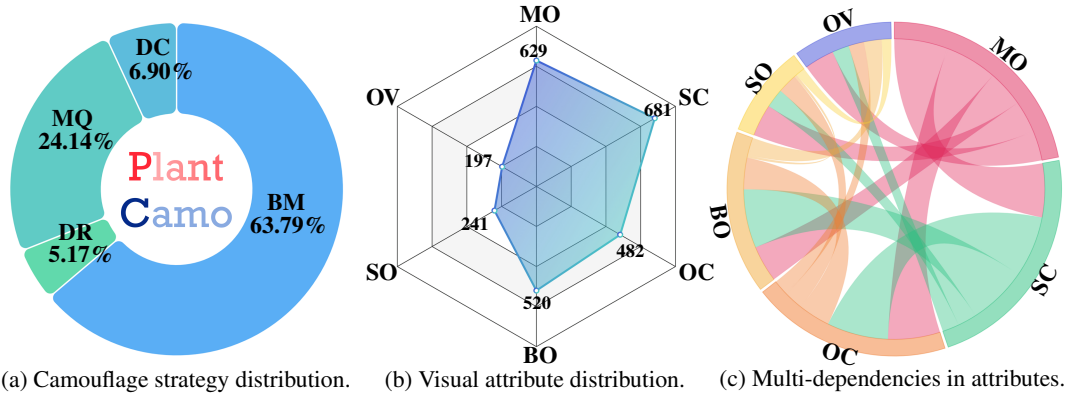

\section{Dataset Construction}\label{dataset}
\subsection{Data Collection}
Compared to animal camouflage and synthetic camouflage, camouflaged plants are even rarer, making the collection of plant camouflage data an exceedingly challenging task. To construct the PlantCamo dataset, we first categorize the types of camouflaged plants that have been studied by biologists. We then search for images related to these plant categories on websites such as Flickr\footnote{\url{https://www.flickr.com/}}, which provides a repository of public-domain stock photos free from copyright and loyalties. Images are selected based on the following criteria:
1) \textbf{Camouflage:} Plants in the image exhibit clear camouflage characteristics.
2) \textbf{Diversity:} The images cover a wide range of camouflage strategies and camouflaged plants.
3) \textbf{Challenge:} The images include various levels of visual difficulties in distinguishing objects.
We will release this dataset, confirming that all images are available for academic use, to enable researchers to explore plant camouflage in various vision tasks.
More details can be found on the dataset page\footnote{\url{https://github.com/yjybuaa/PlantCamo}}.



\subsection{Data Annotation}
Inspired by \cite{cod10kpami}, we hierarchically annotated PlantCamo. 
To be more precise, for each image, we annotated the plant category and image attributes related to camouflage strategy and visual challenges. 
For each camouflaged object in the image, we provided bounding box-, object-, and instance-level mask annotations. 
The annotation process was as follows: \textit{object category $\rightarrow$ camouflage strategy attributes $\rightarrow$ object bounding box $\rightarrow$ object mask $\rightarrow$ vision challenge attributes $\rightarrow$ object instance}. Annotation samples are shown in Fig.~\ref{annotation}.
These rich annotations enabled the PlantCamo dataset to be used for multiple vision tasks. Here, we would like to specifically introduce the attribute annotation.

\textbf{Attributes.}
Unlike prior works, we annotate the attributes hierarchically in both category- and image-level.
For category-level attributes, we annotate each plant category based on the plant camouflage strategies, resulting in background matching (BM), disruptive coloration (DC), masquerade (MQ), and decoration (DR).
The category-level attributes can help us analyze the differences between various camouflage strategies in COD.
For image-level attributes, we select 6 attributes from COD10K~\cite{COD10K}: multiple objects (MO), shape complexity (SC), occlusion (OC), big object (BO), small object (SO) and out-of-view (OV), which are commonly faced but challenging in plant camouflage detection.
Detailed description for each attribute is given in Tab.~\ref{attribute}.

\subsection{Data Statistics}


\textbf{Category distribution.}
In PlantCamo, 58 kinds of camouflaged plants are involved.
The distribution of plant categories is given in Fig.~\ref{fig:category}.
As shown, we also give some representative examples in PlantCamo for visualization.
The content of PlantCamo covers most types that are investigated in research conducted on plant camouflage. 

\textbf{Dataset composition.}
We utilize the PlantCamo dataset in 3 subsets: PlantCamo-full, PlantCamo-train, and PlantCamo-test. The PlantCamo-full set encompasses the entire collection of 1,250 camouflaged plant images, primarily utilized for evaluating the generalization capabilities of COD models. Furthermore, we divide the PlantCamo dataset into two subsets: PlantCamo-train, containing 1,000 images for training purposes, and PlantCamo-test, comprising 250 images designated for testing. 
Note that we carefully choose the test samples, and some categories of test samples do even not appear in the training set, to give a thorough evaluation of model learning ability. 


\textbf{Attribute distribution.}
As we annotate the attributes from two distinct perspectives - camouflage strategies and visual attributes - we present the distribution of each separately for clarity. In Fig.~\ref{fig:subfig11}, the distribution of plant camouflage strategies is clearly depicted. Notably, background matching emerges as the most prevalent strategy, accounting for a substantial 63.79\% of the entire dataset. Subsequently, Figs.~\ref{fig:subfig7} and \ref{fig:subfig8} offer insights into the distributions of visual attributes.

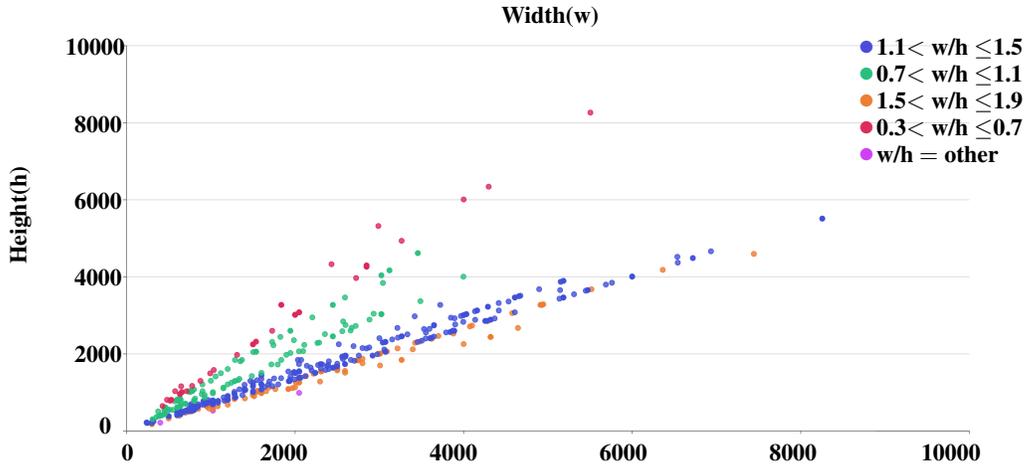
\begin{figure}[t]
  \centering
   \begin{overpic}[width=0.85\linewidth]{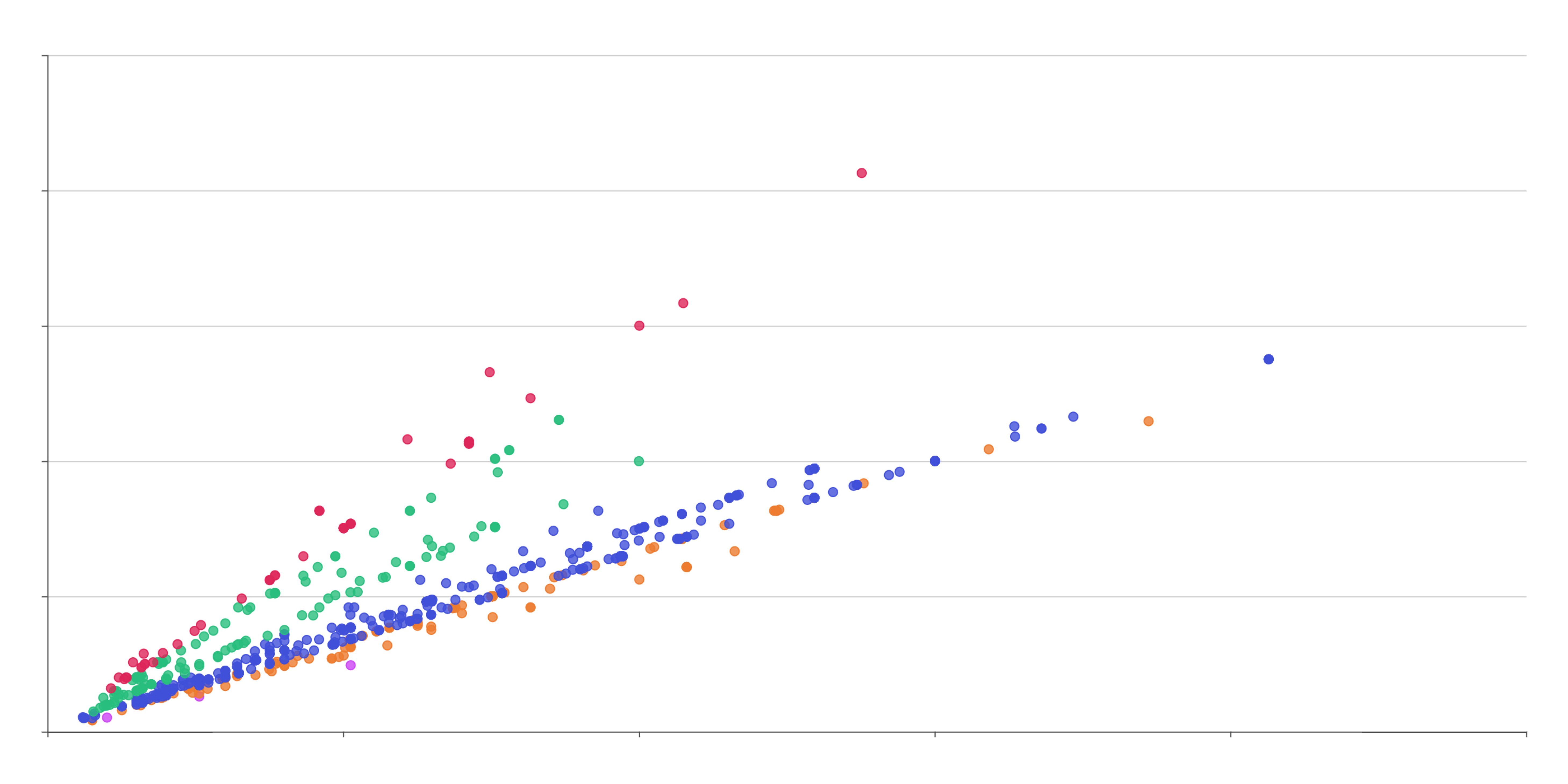}
    \put(85,45.4){\large {\textbf{\color[RGB]{76,75,215}$\bullet$}}}
    \put(87,45.4){\footnotesize {\textbf{1.1$\textless$~w/h~$\leq$1.5}}}
    \put(85,42.4){\large {\textbf{\color[RGB]{40,191,126}$\bullet$}}}
    \put(87,42.4){\footnotesize {\textbf{0.7$\textless$~w/h~$\leq$1.1}}}
    \put(85,39.4){\large {\textbf{\color[RGB]{235,127,52}$\bullet$}}}
    \put(87,39.4){\footnotesize {\textbf{1.5$\textless$~w/h~$\leq$1.9}}}
    \put(85,36.4){\large {\textbf{\color[RGB]{220,45,90}$\bullet$}}}
    \put(87,36.4){\footnotesize {\textbf{0.3$\textless$~w/h~$\leq$0.7}}}
    \put(85,33.4){\large {\textbf{\color[RGB]{206,65,243}$\bullet$}}}
    \put(87,33.4){\footnotesize {\textbf{w/h~$=$~other}}}

    \put(-3.8,45.4){\footnotesize {\textbf{10000}}}
    \put(-2.8,36.8){\footnotesize {\textbf{8000}}}
    \put(-2.8,28.2){\footnotesize {\textbf{6000}}}
    \put(-2.8,19.6){\footnotesize {\textbf{4000}}}
    \put(-2.8,11){\footnotesize {\textbf{2000}}}
    \put(0,3){\footnotesize {\textbf{0}}}
    \put(-10,22){\footnotesize \rotatebox{90} {\textbf{Height(h)}}}

    \put(2.5,0){\footnotesize {\textbf{0}}}
    \put(18,0){\footnotesize {\textbf{2000}}}
    \put(37,0){\footnotesize {\textbf{4000}}}
    \put(56,0){\footnotesize {\textbf{6000}}}
    \put(75,0){\footnotesize {\textbf{8000}}}
    \put(92,0){\footnotesize {\textbf{10000}}}
    \put(45,49){\footnotesize {\textbf{Width(w)}}}   
     \end{overpic}  
  \caption{Image resolution distribution.}
  \label{fig:subfig10}
\end{figure}

\textbf{Image resolution.}
In Fig.~\ref{fig:subfig10}, we present the distribution of resolutions for the proposed dataset. 
Compared to existing datasets~\cite{CAMO,NC4K,cod10kpami}, our PlantCamo contains a significantly larger number of high-resolution images, specifically those with a resolution exceeding $2K\times2K$. 
The inclusion of these high-resolution images enables the capture of finer texture details, which is highly beneficial for camouflage detection~\cite{highreso}, as many COD models rely on boundaries and textures to distinguish camouflaged objects~\cite{boundary1,findnet}.

\begin{table*}[t]
\setlength\tabcolsep{2pt}
\renewcommand{\arraystretch}{1.0}
\footnotesize
\centering
\caption{Quantitative results of SoTA on PlantCamo. 
Top 3 results are shown in \textbf{\color{red}red}, \textbf{\color{blue}blue}, and \textbf{\color[rgb]{0,0.7,0}green}.}
\label{Compare SoTA}
\begin{tabular}{l|r|l|ccccccccc}
\toprule
Method &Venue & Backbone& $S_\alpha\uparrow$  & $F^w_\beta\uparrow$ & $M\downarrow$  & $E^{ad}_\varphi\uparrow$ & $E^{m}_\varphi\uparrow$  & $E^{max}_\varphi\uparrow$ & $F^{ad}_\beta\uparrow$& $F^{m}_\beta\uparrow$& $F^{max}_\beta\uparrow$\\
\midrule
\textbf{SINet}~\cite{COD10K}        &CVPR$_{20}$  &ResNet-50~\cite{ResNet}     &0.608 	&0.410 	&0.127 	&0.681 	&0.627 	&0.725 	&0.518 	&0.478 	&0.530   \\
\textbf{MGL}~\cite{MGL}             &CVPR$_{21}$  &ResNet-50~\cite{ResNet}     &0.591 	&0.364 	&0.130 	&0.691 	&0.574 	&0.721 	&0.500 	&0.419 	&0.499   \\
\textbf{PFNet}~\cite{PFNet}         &CVPR$_{21}$  &ResNet-50~\cite{ResNet}     &0.637 	&0.455 	&0.118 	&0.709 	&0.660 	&0.739 	&0.557 	&0.525 	&0.555     \\ 
\textbf{UGTR}~\cite{UGTR}           &ICCV$_{21}$  &ResNet-50~\cite{ResNet}     &0.621 	&0.410 	&0.121 	&0.725 	&0.612 	&0.740 	&0.546 	&0.463 	&0.532    \\
\textbf{C2FNet}~\cite{C2FNet}       &IJCAI$_{21}$ &Res2Net-50~\cite{Res2Net}    &0.631 	&0.434 	&0.119 	&0.688 	&0.632 	&0.694 	&0.527 	&0.490 	&0.525   \\
\textbf{SINet-V2}~\cite{cod10kpami} &PAMI$_{22}$ &Res2Net-50~\cite{Res2Net}    &0.672 	&0.504 	&0.109 	&\textbf{\color[rgb]{0,0.7,0}0.772} 	&\textbf{\color[rgb]{0,0.7,0}0.722} 	&\textbf{\color[rgb]{0,0.7,0}0.781} 	&0.600 	&0.568 	&0.593   \\
\textbf{C2FNet-V2}~\cite{C2FNet-V2} &CSVT$_{22}$ &Res2Net-50~\cite{Res2Net}    &0.651 	&0.474 	&0.111 	&0.705 	&0.684 	&0.714 	&0.540 	&0.531 	&0.545  \\
\textbf{TPRNet}~\cite{TPRNet}       &TVCJ$_{22}$  &Res2Net-50~\cite{Res2Net}    &0.666 	&0.482 	&0.112 	&0.764 	&0.708 	&0.761 	&0.575 	&0.539 	&0.565  \\
\textbf{BSANet}~\cite{BSANet}       &AAAI$_{22}$  &Res2Net-50~\cite{Res2Net}    &0.644 	&0.453 	&0.113 	&0.712 	&0.645 	&0.739 	&0.561 	&0.510 	&0.555  \\
\textbf{FAPNet}~\cite{FAPNet}       &TIP$_{22}$   &Res2Net-50~\cite{Res2Net}    &0.639 	&0.452 	&0.117 	&0.728 	&0.658 	&0.750 	&0.564 	&0.514 	&0.563  \\
\textbf{BGNet}~\cite{BGNet}         &IJCAI$_{22}$ &Res2Net-50~\cite{Res2Net}    &0.625 	&0.429 	&0.123 	&0.691 	&0.634 	&0.702 	&0.523 	&0.484 	&0.531  \\
\textbf{SegMaR}~\cite{SegMaR}       &CVPR$_{22}$  &ResNet-50~\cite{ResNet}     &0.545 	&0.308 	&0.156 	&0.696 	&0.546 	&0.750 	&0.515 	&0.381 	&0.547  \\
\textbf{ERRNet}~\cite{ERRNet}       &PR$_{22}$    &ResNet-50~\cite{ResNet}     &0.626 	&0.402 	&0.132 	&0.743 	&0.636 	&0.740 	&0.546 	&0.470 	&0.528   \\
\textbf{OCENet}~\cite{OCENet}       &WACV$_{22}$  &ResNet-50~\cite{ResNet}     &0.645 	&0.459 	&0.113 	&0.718 	&0.665 	&0.738 	&0.560 	&0.519 	&0.555  \\
\textbf{PreyNet}~\cite{PreyNet}     &MM$_{22}$ &ResNet-50~\cite{ResNet}     &0.609 	&0.400 	&0.122 	&0.668 	&0.602 	&0.706 	&0.506 	&0.460 	&0.514   \\
\textbf{ZoomNet}~\cite{ZoomNet}     &CVPR$_{22}$  &ResNet-50~\cite{ResNet}     &0.635	&0.430 	&0.111 	&0.708 	&0.619 	&0.725 	&0.547 	&0.477 	&0.541  \\
\textbf{DGNet}~\cite{DGNet}         &MIR$_{23}$   &EffNet-B4~\cite{Eff}     &0.657 	&0.480 	&0.114 	&0.736 	&0.683 	&0.760 	&0.582 	&0.543 	&0.579  \\
\textbf{DTINet}~\cite{DTINet}       &ICPR$_{22}$  &MiT-B5~\cite{MiT}        &\textbf{\color{blue}0.706} 	&\textbf{\color{blue}0.551} 	&\textbf{\color{blue}0.099} 	&\textbf{\color{blue}0.785} 	&\textbf{\color{blue}0.754} 	&\textbf{\color{blue}0.789} 	&0.615 	&\textbf{\color{blue}0.597} 	&0.608  \\
\textbf{FSPNet}~\cite{FSPNet}       &CVPR$_{23}$  &ViT~\cite{ViT}           &0.466 	&0.158 	&0.168 	&0.703 	&0.369 	&0.693 	&0.476 	&0.175 	&0.480  \\
\textbf{DaCOD}~\cite{dacod}              &MM$_{23}$ &Hybrid~\cite{ResNet}\cite{Swin}        &0.671 	&0.507 	&0.110 	&0.741 	&0.682 	&0.773 	&\textbf{\color[rgb]{0,0.7,0}0.622} 	&0.573 	&\textbf{\color{blue}0.624}  \\
\textbf{HitNet}~\cite{HitNet}       &AAAI$_{23}$  &PVTv2-B2~\cite{pvtv2} &\textbf{\color{red}0.740} 	&\textbf{\color{red}0.618} 	&\textbf{\color{red}0.081} 	&\textbf{\color{red}0.795} 	&\textbf{\color{red}0.784} 	&\textbf{\color{red}0.803} 	&\textbf{\color{red}0.664} 	&\textbf{\color{red}0.657} 	&\textbf{\color{red}0.666}  \\
\textbf{VSCode}~\cite{vscode}       &CVPR$_{24}$  &Swin-T~\cite{Swin} &\textbf{\color[rgb]{0,0.7,0}0.698} & \textbf{\color[rgb]{0,0.7,0}0.547} & \textbf{\color[rgb]{0,0.7,0}0.102} & 0.766 & \textbf{\color[rgb]{0,0.7,0}0.722} & 0.778 & \textbf{\color{blue}0.628} & \textbf{\color[rgb]{0,0.7,0}0.595} & 0.619 \\
\textbf{CamoDiffusion}~\cite{camodiffusion}       &AAAI$_{24}$  &PVTv2-B4~\cite{pvtv2} &0.682 & 0.516 & \textbf{\color[rgb]{0,0.7,0}0.102} & 0.706 & 0.699 & 0.706 & 0.561 & 0.556 & 0.561\\
\textbf{CamoFormer}~\cite{camoformer}       &PAMI$_{24}$  &PVTv2-B4~\cite{pvtv2}&0.681 &0.522 &0.106 &0.747 &0.695 &0.776 &\textbf{\color[rgb]{0,0.7,0}0.622} &0.574 &\textbf{\color[rgb]{0,0.7,0}0.620}\\
\bottomrule
\end{tabular}
\end{table*}

\section{Benchmark Study}\label{bench}

\subsection{Benchmark Settings}

\textbf{Tested models.} To provide a thorough benchmark study on plant camouflage detection, we evaluate the capabilities of current cutting-edge COD models on the proposed PlantCamo.
Regarding the code availability, we here choose 24 state-of-the-art models for comparison: 
SINet~\cite{COD10K}, C2FNet~\cite{C2FNet}, MGL~\cite{MGL}, PFNet~\cite{PFNet}, UGTR~\cite{UGTR}, SINet-V2~\cite{cod10kpami}, C2FNet-V2~\cite{C2FNet-V2}, ERRNet~\cite{ERRNet}, TPRNet~\cite{TPRNet}, BSANet~\cite{BSANet}, FAPNet~\cite{FAPNet}, DTINet~\cite{DTINet}, BGNet~\cite{BGNet}, OCENet~\cite{OCENet}, PreyNet~\cite{PreyNet}, ZoomNet~\cite{ZoomNet}, SegMaR~\cite{SegMaR}, FSPNet~\cite{FSPNet}, DGNet~\cite{DGNet}, DaCOD~\cite{dacod}, HitNet~\cite{HitNet}, VSCode~\cite{vscode}, CamoDiffusion~\cite{camodiffusion}, and CamoFormer~\cite{camoformer}.
Here, to assess the overall generalization capability of advanced models, we conduct tests using PlantCamo-full.

\textbf{Evaluation metrics.}
For quantitative evaluation, we use 9 standard metrics:
Structure-measure ($S_\alpha$)~\cite{S-measure}, adaptive E-Measure ($E^{ad}_\varphi$)~\cite{E-measure}, mean E-Measure ($E^{m}_\varphi$), max E-Measure ($E^{max}_\varphi$), adaptive F-Measure ($F^{ad}_\beta$)~\cite{F-measure-mean}, mean F-Measure ($F^{m}_\beta$), max F-Measure ($F^{max}_\beta$), Weighted F-Measure ($F_\beta^w$)~\cite{F-measure}, and Mean Absolute Error ($M$).

\subsection{Benchmarking SoTA COD Models}
Quantitative results on PlantCamo are given in Tab.~\ref{Compare SoTA}.
Obviously, HitNet~\cite{HitNet}, which is a recent Transformer-based method, achieves the best performance on PlantCamo.
It surprisingly surpasses the second DTINet~\cite{DTINet} by a large margin, \textit{i.e.}, 3.4\% in terms of $S_\alpha$, thanks to its ability to extract high-resolution texture details, which helps on the camouflage detection on high-quality PlantCamo images.
DTINet~\cite{DTINet} also performs well and occupies the second place, while note that DTINet uses a heavy Transformer and has parameters of 266.33M~\cite{advances}.
We also notice that the overall performance on PlantCamo is much lower than the ones on previous camouflage datasets~\cite{cod10kpami, NC4K}, which shows that the plant camouflage is very challenging to existing COD models.

\begin{figure}
  \centering
  \includegraphics[width=.95\linewidth]{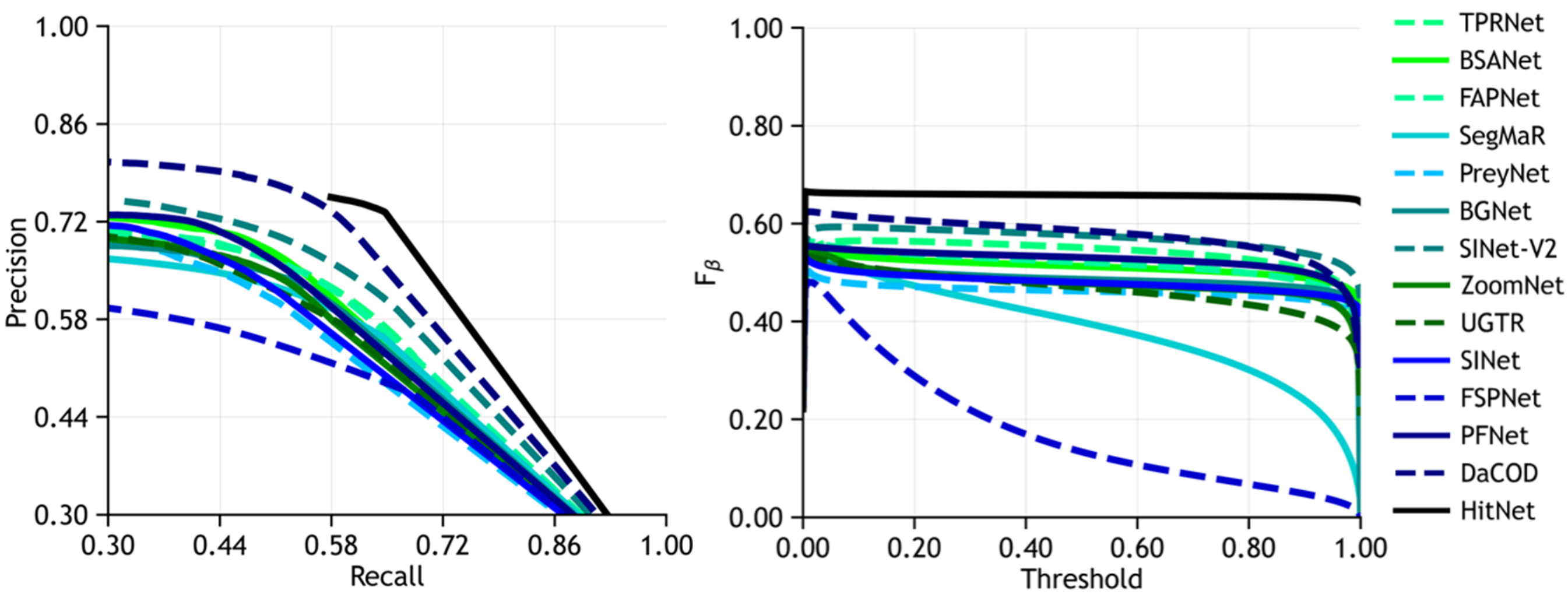}
  \caption{PR and $F_{\beta}$ curves of the recent SoTA algorithms on the PlantCamo dataset.}
    \label{pr_fm}
\end{figure}

We also present the precision-recall (PR) and $f_{\beta}$ curves of previous methods, on the PlantCamo dataset, as shown in Fig.~\ref{pr_fm}. Note that a higher curve indicates better model performance.
Obviously, HitNet~\cite{HitNet}, which is a recent Transformer-based method, achieves the best performance on PlantCamo.
However, compared with other COD benchmarks, the overall performance of current SoTAs is still unsatisfactory on PlantCamo.

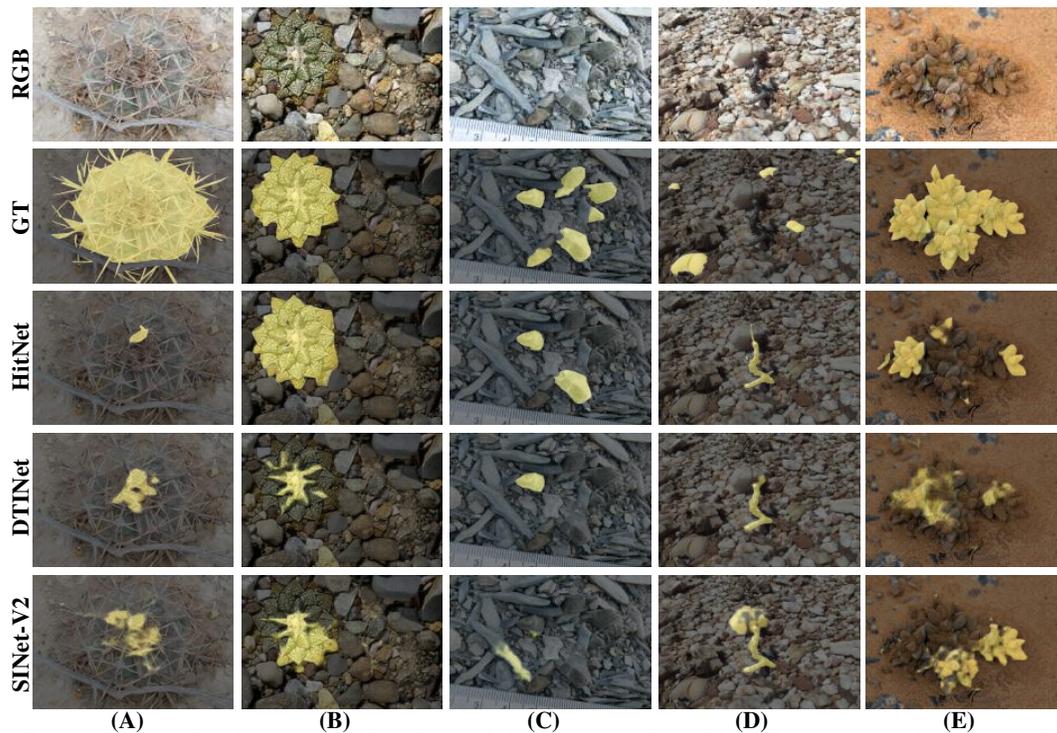
\begin{figure}
	\centering
    \small
 \begin{overpic}[width=0.99\linewidth]
{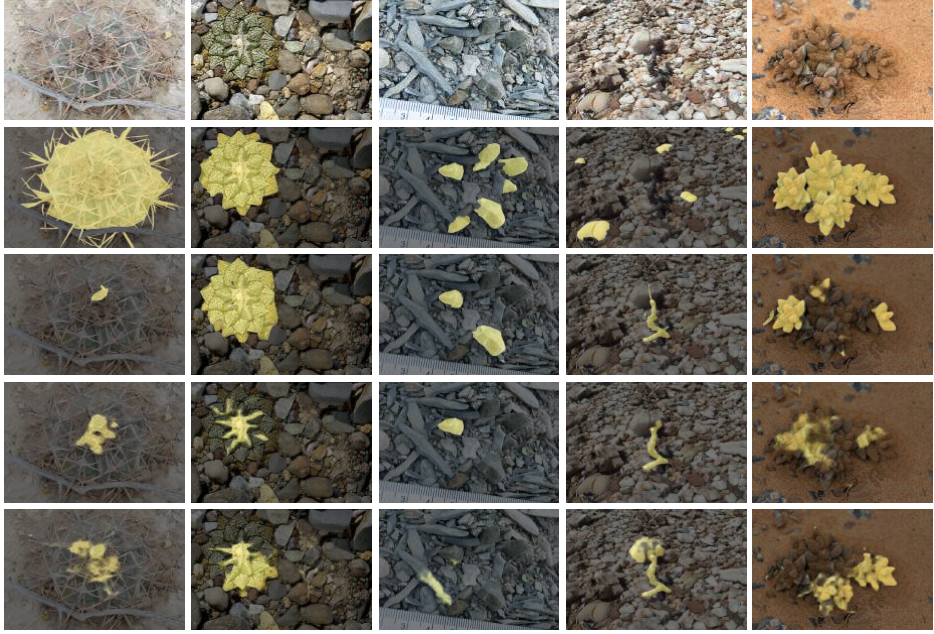}
    \put(8,-1.5){\footnotesize \textbf{(A)}}
    \put(28,-1.5){\footnotesize \textbf{(B)}}
    \put(48,-1.5){\footnotesize \textbf{(C)}}
    \put(68,-1.5){\footnotesize \textbf{(D)}}
    \put(88,-1.5){\footnotesize \textbf{(E)}}
    \put(-1.5,59){\footnotesize \rotatebox{90}{\textbf{RGB}}} 
    \put(-1.5,46){\footnotesize \rotatebox{90}{\textbf{GT}}}
    \put(-1.5,31){\footnotesize \rotatebox{90}{\textbf{HitNet}}}
    \put(-1.5,17){\footnotesize \rotatebox{90}{\textbf{DTINet}}}
    \put(-1.5,2){\footnotesize \rotatebox{90}{\textbf{SINet-V2}}}
    \end{overpic}  
	\caption{Visualized results in PlantCamo. 
    We show samples with different camouflage strategies, \textit{i.e.}, BM (column A), DC (column B), MQ (column C-D), and DR (column E).}
 \label{benchmark2}
\end{figure}

\textbf{Visualized results.}
We further give visualized results of the models for analysis, in which different camouflage strategies are covered.
As shown in Fig.~\ref{benchmark2}, even current SoTAs are fooled by the plant camouflage strategies, suffering from wrong boundary descriptions and missed detection.
Obviously, in cases (A) and (B), current advanced COD models cannot distinguish and segment the camouflaged plants precisely.
Meanwhile, for (C) to (E), some of the camouflaged plants are missed by the COD models.
Thus, plant camouflage is difficult to distinguish by current COD models, which leaves large room for further improvement.


\subsection{Discussion}\label{discussion}

Based on the experiments, we here discuss the differences between animal and plant camouflage, and difficulties in plant camouflage, to analyze the failures of current SoTA and inspire new ideas.

\textbf{Differences with animal camouflage.}
Animal camouflage and plant camouflage are both concealment methods to help living beings adapt to their environments. 
Animal camouflage typically involves the use of coloration, texture, and shape to blend into the background and avoid detection by predators.
Plant camouflage, however, typically involves the use of coloration and shape to deceive animals into thinking they are something else. 
In addition to these differences, animals have more flexibility in terms of movement and behavior, allowing them to actively seek out prey or predators and avoid danger. Plants, on the other hand, are sessile organisms that heavily rely on external factors like wind, animals, and other means to disperse their seeds and propagate. 
Thus, plant camouflage is more geared towards deception to ensure their survival and reproduction.

\textbf{Challenges in PCD.}
Our benchmark study reveals several challenges for current COD models in PCD. 
One of the primary difficulties lies in the unique characteristics of camouflaged plants compared to animals. Unlike animals that camouflage by blending into their backgrounds, plants often disguise themselves as stones or other lifeless objects. This often leads to plants being scattered in various positions within the field of view, making it challenging for COD models to accurately detect them.
Furthermore, the edges of camouflaged plants are often complex and irregular, sometimes even featuring burrs or other vegetative protuberances. This can significantly hinder the accurate segmentation of plant shapes and boundaries. As a result, even advanced COD models struggle to produce precise detections, leading to potential missed detections or false positives.
These challenges emphasize the need for further research on PCD to address these unique difficulties effectively.

\begin{figure*}[t]
\centering
  \includegraphics[width=0.99\linewidth]{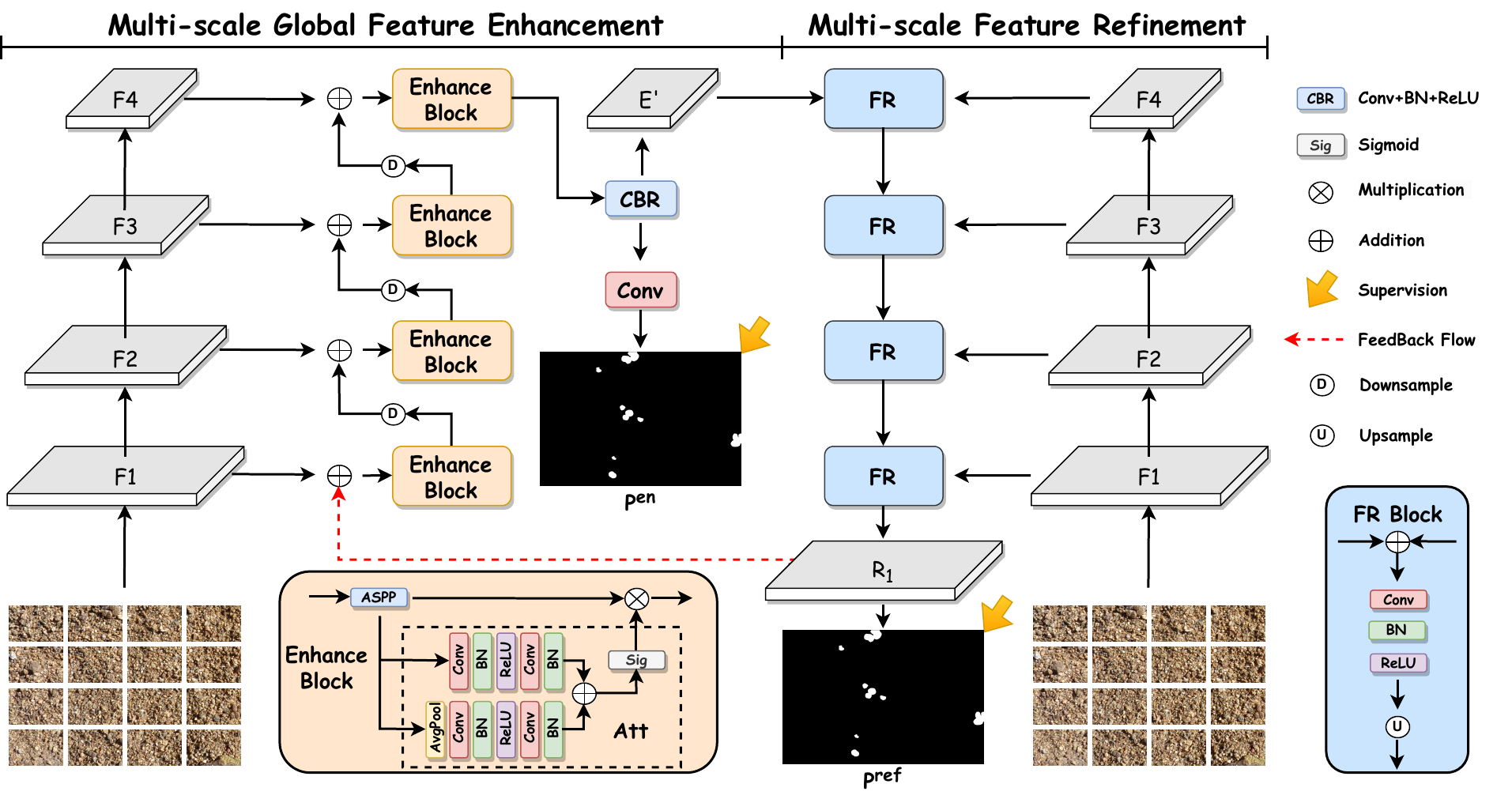}
  \caption{The overall architecture of the PCNet, which consists of two key components: Multi-scale Global Feature Enhancement (MGFE) module and Multi-scale Feature Refinement (MFR) module.}
  \label{Overview}
\end{figure*}

\section{Method}
Inspired by the plant-specific characteristics discussed above, we newly propose PCNet, aiming to solve the difficulties in PCD.
As illustrated in Fig.~\ref{Overview}, PCNet consists of two main parts: Multi-scale Global Feature Enhancement (MGFE) and Multi-scale Feature Refinement (MFR) modules. To detect camouflaged plants accurately, MGFE enhances and merges features of various scales using a bottom-up approach, providing globally enhanced features. To segment plants with irregular borders and similar shapes to the background, MFR refines and merges features at different scales using a top-down approach with iterative feedback,
which results in precise segmentation. Together, MGFE and MFR contribute to accurate PCD in complex environments.

\subsection{Multi-scale Global Feature Enhancement}
PVT~\cite{pvtv2} is famous as the backbone network due to its capability to extract multi-scale features and provide relatively higher-resolution feature maps.
Therefore, we chose PVT as the feature extractor to facilitate more effective extraction of high-resolution feature maps without excessively taxing memory resources.
Then, we extract features of different scales, denoted as $\{F_i~|~i=1,2,3,4\}$, and input them sequentially in descending order into the enhancement module to obtain enhanced features, denoted as $\{E_i~|~i=1,2,3,4\}$. 
We perform feature enhancement and fusion across different scales in a bottom-up manner. The enhancement module~\cite{MENet} consists of ASPP~\cite{ASPP} and attention block. 
This can be specified more precisely by the following equations:
\begin{gather}
     E_i \!= \!ASPP(F_i)\cdot Att(ASPP(F_i)),  i\!=\!2,3,4  \\
     E_1\! =\! ASPP(F_1\!+\!B_j)\cdot Att(ASPP(F_1\!+\!B_j)), j\!=\!1,2 
\end{gather}
where $Att$ represents the attention module, and $B_j$ represents the feedback features after the j-th iteration, where $B_1 = 0$ when $j$ equals 1. 
Next, $E_4$ is processed through the CBR (Conv+Batch Normalization+ReLU) operation to obtain the final enhanced feature $E'$, and an initial prediction map $P^{en}$ is obtained by applying a $7\times7$ convolution. 
The process can be formulated as follows:
\begin{gather}
    E' = Conv(BN(ReLU(E_4))), \\
    P^{en} = Conv(E').
\end{gather}

\begin{table*}[t]
\setlength\tabcolsep{4pt}
\scriptsize
\centering
\caption{Quantitative results on the proposed PlantCamo-test set. Here ``w/o" and ``w" indicate the model is trained on the PlantCamo-train set.
Top 3 results are shown in {\color{red}red}, {\color{blue}blue}, and {\color[rgb]{0,0.7,0}green}.}
\label{testcomparison}
\begin{tabular}{r cc cc cc cc cc cc cc c}
\toprule
\multirow{2}{*}{Method} 
& \multicolumn{2}{c}{\textbf{ZoomNet}~\cite{ZoomNet}}& \multicolumn{2}{c}{\textbf{SINet-V}2~\cite{cod10kpami}}& \multicolumn{2}{c}{\textbf{PFNet}~\cite{PFNet}}
&\multicolumn{2}{c}{\textbf{SINet}~\cite{COD10K}}& \multicolumn{2}{c}{\textbf{FSPNet}~\cite{FSPNet}}&\multicolumn{2}{c}     {\textbf{DaCOD}~\cite{dacod}}&\multicolumn{2}{c}{\textbf{HitNet}~\cite{HitNet}} &\multicolumn{1}{c}{\textbf{Ours}}\\
&w/o & w &w/o & w &w/o & w  &w/o & w  &w/o & w  &w/o & w &w/o & w  & w  \\
\cmidrule(lr){1-1}\cmidrule(lr){2-3}\cmidrule(lr){4-5}\cmidrule(lr){6-7}\cmidrule(lr){8-9}\cmidrule(lr){10-11}\cmidrule(lr){12-13}\cmidrule(lr){14-15}\cmidrule(lr){16-16}
$S_\alpha \uparrow$            
 &0.560&0.798   &0.593&0.801   &0.567&0.787     &0.540&0.766 &0.453&\textbf{\color[rgb]{0,0.7,0}0.816}&0.584&0.804&0.637&\textbf{\color{blue}0.854}   &\textbf{\color{red}0.880} \\
$F^w_\beta \uparrow$            
 &0.281&0.680   &0.355&0.678   &0.313&0.660     &0.271&0.583 &0.093&\textbf{\color[rgb]{0,0.7,0}0.703}&0.334&0.693  &0.436&\textbf{\color{blue}0.794} &\textbf{\color{red}0.818} \\
$M \downarrow$                   
 &0.111&0.049   &0.112&0.050   &0.114&0.054     &0.121&0.066 &0.143&\textbf{\color[rgb]{0,0.7,0}0.042}&0.109&0.046  &0.102&\textbf{\color{blue}0.034}&\textbf{\color{red}0.028} \\
$E^{ad}_\varphi \uparrow$          
 &0.650&0.874   &0.721&0.873   &0.665&0.868     &0.640&0.842 &0.675&0.876&0.659&\textbf{\color[rgb]{0,0.7,0}0.889 } &0.708&\textbf{\color{blue}0.929}&\textbf{\color{red}0.937} \\
\midrule
\midrule
\multirow{2}{*}{Method}
& \multicolumn{2}{c}{\textbf{UGTR}~\cite{UGTR}}& \multicolumn{2}{c}{\textbf{BGNet}~\cite{BGNet}}& \multicolumn{2}{c}{\textbf{PreyNet}~\cite{PreyNet}}
&\multicolumn{2}{c}{\textbf{SegMaR}~\cite{SegMaR}}& \multicolumn{2}{c}{\textbf{FAPNet}~\cite{FAPNet}}&\multicolumn{2}{c}     {\textbf{BSANet}~\cite{BSANet}}&\multicolumn{2}{c}{\textbf{TPRNet}~\cite{TPRNet}} &\multicolumn{1}{c}{\textbf{Ours}}\\
&w/o & w &w/o & w &w/o & w  &w/o & w  &w/o & w  &w/o & w &w/o & w  & w  \\
\cmidrule(lr){1-1}\cmidrule(lr){2-3}\cmidrule(lr){4-5}\cmidrule(lr){6-7}\cmidrule(lr){8-9}\cmidrule(lr){10-11}\cmidrule(lr){12-13}\cmidrule(lr){14-15}\cmidrule(lr){16-16}
$S_\alpha \uparrow$            
&0.565&0.804 &0.559&0.786&0.543&0.791 &0.522&0.791 &0.570&0.801&0.563&0.782&0.588&\textbf{\color[rgb]{0,0.7,0}0.816}&\textbf{\color{red}0.880} \\
$F^w_\beta \uparrow$            
 &0.296&0.668 &0.293&0.552&0.254&0.686 &0.233&0.657 &0.311&0.684&0.298&0.663&0.336&0.680&\textbf{\color{red}0.818} \\
$M \downarrow$                   
 &0.115&0.050 &0.121&0.076 &0.117&0.048 &0.133&0.055 &0.113&0.051&0.112&0.053&0.116&0.051&\textbf{\color{red}0.028} \\
$E^{ad}_\varphi \uparrow$          
&0.693&0.862 &0.648&0.873 &0.616&0.883 &0.688&0.855 &0.690&0.879&0.653&0.875&0.706&0.872&\textbf{\color{red}0.937} \\
 \bottomrule
\end{tabular}
\end{table*}

\subsection{Multi-scale Feature Refinement}

The right part in Fig.~\ref{Overview} displays our Multi-scale Feature Refinement (MFR) module. 
The enhanced features $E'$ from the MGFE module and the features $F_4$ extracted by PVT~\cite{pvtv2} are input into the top $F\!R$ Block for feature refinement. 
Through the top-down fusion process, finer features $\{R_{i~|~i=4,3,2,1}\}$ are continuously restored. 
This process can be represented as follows:
    \begin{gather}
    R_4 = F\!R(E'+F_4),\\
    R_i = F\!R(R_{i+1}+F_i) , i=3,2,1
    \end{gather}
where $F\!R$ represents the FR Block, and $F_i$ represents the multi-level features extracted by PVT.
The feature refinement process occurs from top to bottom, resulting in a more refined feature $R_1$ and prediction map $P^{ref}$ in the end.

\textbf{Refinement by iterative feedback.}
The edges of camouflaged plants are often highly intricate and irregular, making them difficult to be recognized and segmented accurately. 
To address this challenge, we introduce an iterative feedback strategy that is essential for capturing the more fine-grained details of these plants.
Following the extraction of the refined feature $R_1$, it is used as the feedback feature $B_j$, as shown in Fig.~\ref{Overview}. 
Thus, the feedback feature can be expressed as $B_j=R_1^j$, where $j$ represents the $j$-th iteration. 
It is noteworthy that when $j$ equals 1, $B_1 = 0$.
This feedback feature is then fused with the initial feature $F_1$, and fed back into the MGFE module.
We believe that the globally enhanced features obtained through bottom-up fusion and the refined features from top-down fusion complement each other, leading to a more comprehensive understanding of plant camouflage patterns.

\subsection{Loss Function}

Our loss is divided into two parts: one is for the MGFE module, and the other one is for the MFR module. 
Thus, we use a combination of weighted binary cross-entropy loss (BCE) and weighted IoU loss, which can be represented as $L_{base} = l_{wbce}+l_{wiou}$.
As we employ an iterative feedback mechanism, losses in later iterations should be more significant. Therefore, similar to the settings in HitNet~\cite{HitNet}, we define a weight parameter $\mu$. 
The losses generated by the MGFE and MFR modules can be represented as follows:
\begin{gather}
    L_{e} =\sum\limits_{i=1}^{2} ((i-1)\cdot\mu) \cdot L_{base}(P^{{en}^{i}},GT);\\
    L_{r} =\sum\limits_{i=1}^2 ((i-1)\cdot\mu) \cdot L_{base}(P_{1}^{{ref}^{i}},GT),
\end{gather}
where $i$ represents the current iteration number, and $GT$ denotes the ground truth image. Overall loss function can be expressed as: $L_{final} = L_e + L_r$.

\section{Experiments}\label{exp}
\subsection{Experimental Setup}
To validate the proposed PCNet, which is specifically designed for plant camouflage, and make a fair comparison, we here use PlantCamo-train for training and PlantCamo-test for evaluation.
We implement our model based on PyTorch in a 32GB NVIDIA Tesla V100. 
We train our model with input sizes of $704\times704$ and employ PVT-V2~\cite{pvtv2} for feature extraction, as it can capture multi-scale features and provide relatively higher-resolution feature maps. 
The learning rate is set to 1$e$-4, with a decay rate of 0.1. We use the AdamW optimizer with a batch size of 8 and trained for 150 epochs. 
The weight parameter is set as $\mu=0.2$. 
During testing, we used $704\times704$ as the input size, and the output was subsequently restored to the original image dimensions for evaluation. 
For iteration times, we use $j=2$ empirically.


\subsection{Experimental Results}

\begin{figure}
  \centering
  \includegraphics[width=\linewidth]{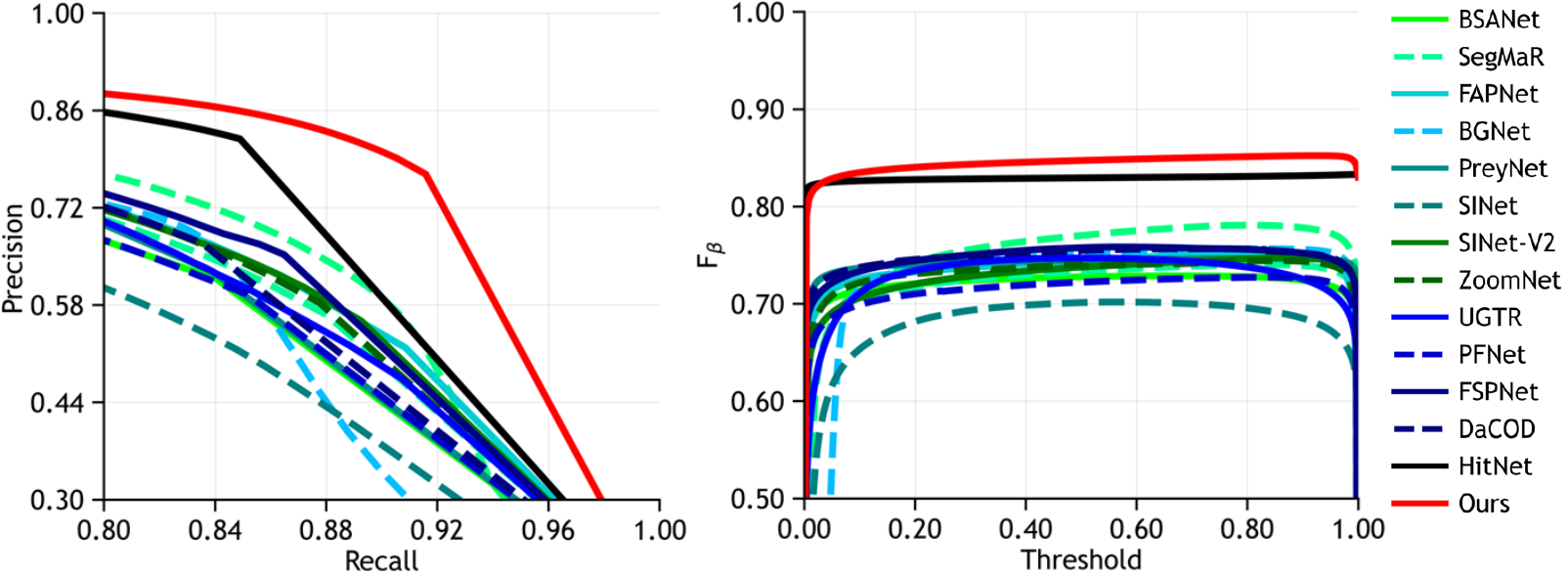}
  \caption{PR and $F_{\beta}$ curves of the proposed PCNet and the SoTA algorithms on PlantCamo-test.}
  \label{fig:subfig250}
\end{figure}

\textbf{Quantitative comparison.}
As presented in Tab.~\ref{testcomparison}, our proposed PCNet tailored for plant camouflage consistently outperformed other techniques by a substantial margin across all metrics. 
When benchmarking against the leading performer, HitNet~\cite{HitNet}, our method showcased a noteworthy average improvement of 6.14\% across all the four key performance indicators.
Compared to FSPNet~\cite{FSPNet}, DaCOD~\cite{dacod}, and TPRNet~\cite{TPRNet}, our method demonstrated an average improvement of 16.12\%, 18.01\%, and 20.17\%, respectively, across all the four metrics.
This indicates that our methods, tailored for PCD, are more effective, compared to approaches designed for camouflaged animals.
In Fig.~\ref{fig:subfig250}, we plotted the PR curves and $F_{\beta}$ curves of our method and existing state-of-the-art methods on the PlantCamo dataset.
Note that the higher the curve is, the better the model performs. 
It is evident that our PCNet outperforms all other methods. 
This observation highlights the effectiveness of PCNet in addressing the challenges associated with plant camouflage.

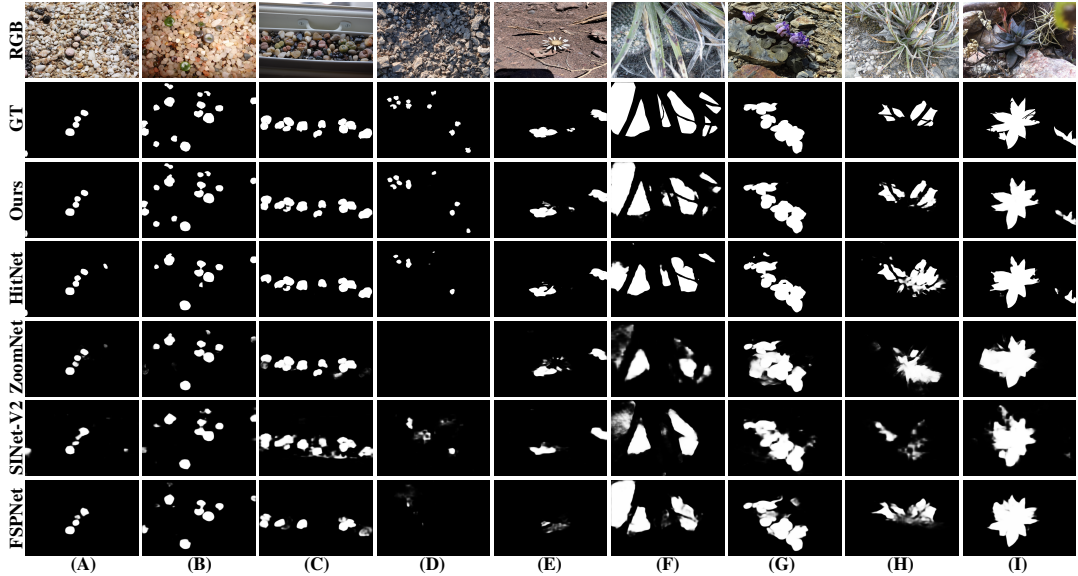
\begin{figure*}[t]
	\centering
    \small
	\begin{overpic}[width=\linewidth]{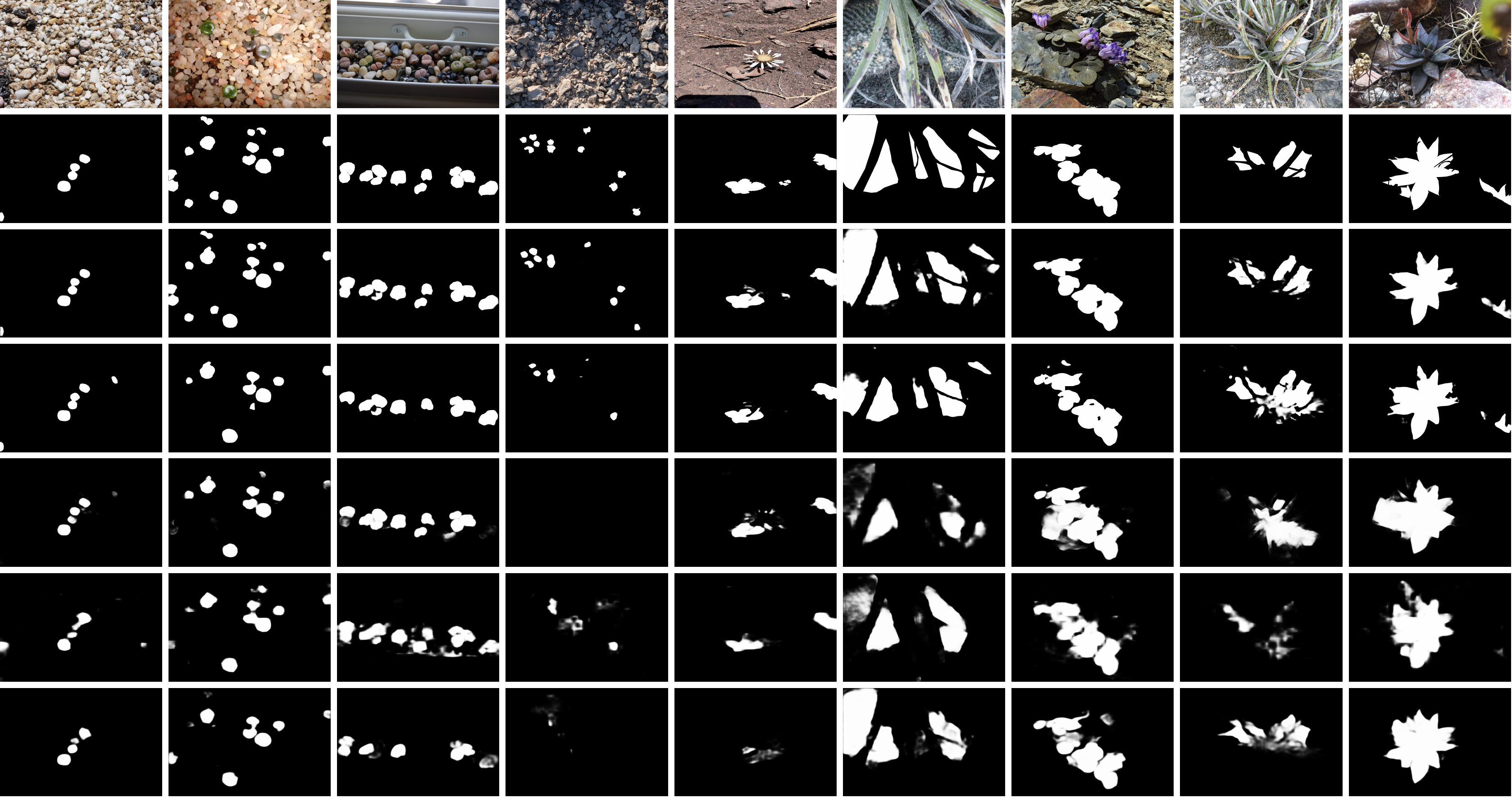} 
    
    \put(4.3,-1.2){\scriptsize \textbf{(A)}}
    \put(15.4,-1.2){\scriptsize \textbf{(B)}}
    \put(26.5,-1.2){\scriptsize \textbf{(C)}}
    \put(37.5,-1.2){\scriptsize \textbf{(D)}}
    \put(48.7,-1.2){\scriptsize \textbf{(E)}}
    \put(60,-1.2){\scriptsize \textbf{(F)}}
    \put(70.7,-1.2){\scriptsize \textbf{(G)}}
    \put(82,-1.2){\scriptsize \textbf{(H)}}
    \put(93.5,-1.2){\scriptsize \textbf{(I)}}
    
    \put(-1.5,47.3){\scriptsize \rotatebox{90}{\textbf{RGB}}} 
    \put(-1.5,40.5){\scriptsize \rotatebox{90}{\textbf{GT}}}
    \put(-1.5,32){\scriptsize \rotatebox{90}{\textbf{Ours}}}
    \put(-1.5,24){\scriptsize \rotatebox{90}{\textbf{HitNet}}}
    \put(-1.5,15.5){\scriptsize \rotatebox{90}{\textbf{ZoomNet}}}
    \put(-1.5,7.8){\scriptsize \rotatebox{90}{\textbf{SINet-V2}}}
    \put(-1.5,0.8){\scriptsize \rotatebox{90}{\textbf{FSPNet}}}

    \end{overpic}
    \vspace{-5pt}
	\caption{Visual comparison with some representative state-of-the-art COD models. 
  Compared to other methods, our approach can produce more accurate results. 
  Please zoom in to see more details.}
 \label{keshihua}
\end{figure*}

\textbf{Qualitative comparison.}
Fig.~\ref{keshihua} provides a visual comparison of our model's results against four other SoTA methods.
As shown in rows (A) to (E), challenges are posed by disguised plants, which are often scattered across various locations and blend in with the background, making them difficult to separate. 
However, our MGFE module effectively addresses these challenges, accurately segmenting the camouflaged plants. In contrast, methods designed for camouflaged animal detection struggle in such scenarios.
Furthermore, rows (F) to (I) demonstrate the challenges posed by camouflaged plants with irregular edges due to occlusions. 
The MFR module plays a crucial role in separating the foreground and background, leading to more precise segmentation of these intricate edges. 
By incorporating an iterative feedback strategy, we are able to achieve even finer boundary separations and reduce background noise interference, compared to other methods.
Overall, our approach demonstrates superior performance.

\subsection{Ablation Study}

\begin{table}[t]
    \centering
    \small
     \renewcommand{\arraystretch}{1.0}
  \setlength\tabcolsep{4pt}
    \caption{Quantitative evaluation for ablation studies.
    ``EB" is Enhance Block, ``FR" is FR Block, and ``FB" is feedback operation.}
    \label{ablation}
    \begin{center}
    \begin{tabular}{ccc|ccccc}
    \toprule
    EB &FR &FB &$S_\alpha \uparrow$&$F^w_\beta \uparrow$&$M \downarrow$&$E^{m}_\varphi \uparrow$&$F^{m}_\beta \uparrow$  \\
    \midrule
        \color[RGB]{128,128,128} &\color[RGB]{128,128,128} &\color[RGB]{128,128,128} &0.866&0.800&0.032&0.922&0.832   \\
        \color[RGB]{128,128,128} &\color[RGB]{128,128,128} &$\checkmark$&0.869&0.804&0.032&0.925&0.837   \\
        \color[RGB]{128,128,128} &$\checkmark$&\color[RGB]{128,128,128} &0.870&0.804&0.031&0.930&0.837   \\
        $\checkmark$&\color[RGB]{128,128,128} &\color[RGB]{128,128,128} &0.869&0.806&0.031&0.925&0.838   \\
        \midrule
        \color[RGB]{128,128,128} &$\checkmark$&$\checkmark$&0.871&0.814&0.030&0.934&0.842   \\
        $\checkmark$&\color[RGB]{128,128,128} &$\checkmark$&0.870&0.810&0.032&0.929&0.846  \\
        $\checkmark$&$\checkmark$&\color[RGB]{128,128,128} &0.877&0.814&0.030&0.933&0.845   \\
        \midrule
$\checkmark$&$\checkmark$&$\checkmark$&\textbf{0.880}&\textbf{0.818}&\textbf{0.028}&\textbf{0.939}&\textbf{0.849}   \\  
    \bottomrule
    \end{tabular}        
    \end{center}
\end{table}

\textbf{Impact of key components.}
We conduct ablation studies by removing different components, \textit{i.e.}, enhance block, FR block, and feedback operation, from our full model.
Corresponding results are given in Tab.~\ref{ablation}.
From the results, we can see the three components all contribute impressively to the model's performance, demonstrating the effectiveness of the proposed modules.
The optimal performance is achieved when all three components are involved together.
Meanwhile, although our method exhibits superior performance, it falls considerably short of the theoretical limits, indicating that more transformative paradigms may be necessary to achieve future breakthroughs in this field.

\begin{table}[t]
\small
    \centering
    \renewcommand{\arraystretch}{1.0}
      \setlength\tabcolsep{3pt}
    \caption{Ablation study on different iteration numbers.}
    \label{iter}
    \begin{center}
    \begin{tabular}{c|ccccc}
    \toprule
    Iteration times   &$S_\alpha \uparrow$  & $F^w_\beta \uparrow$ & $M \downarrow$  & $E^{ad}_\varphi \uparrow$ & $F^{m}_\beta \uparrow$\\
    \midrule
    iter = 1     &0.870 	&0.809 	&0.031 	&0.927 &0.840 	\\
    iter = 2     &\textbf{0.880} 	&\textbf{0.818} 	&\textbf{0.028} 	&\textbf{0.937} &\textbf{0.849} 		\\ 
    iter = 3     &0.872 	&0.811 	&0.030 	&0.934 &0.841 	 \\ 
    iter = 4     &0.876 	&0.816 	&0.030 	&\textbf{0.937} &0.847 	\\ 
    iter = 5     &0.871 	&0.809 	&0.031 	&0.933 	&0.840 	\\ 
    \bottomrule
    \end{tabular}
    \end{center}
\end{table}

\begin{figure}[t]
	\centering
    \small
    \begin{overpic}[width=0.8\linewidth]{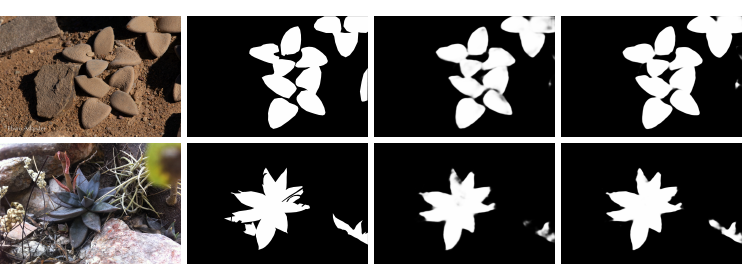}    
    \put(10,-1.5){\scriptsize \textbf{$RGB$}}   
    \put(36,-1.5){\scriptsize \textbf{$GT$}}
    \put(61,-1.5){\scriptsize \textbf{$P_1^{ref}$}}
    \put(86,-1.5){\scriptsize \textbf{$P_2^{ref}$}}
    \end{overpic}
	\caption{Visualized examples of the $P^{ref}$ in each iteration.}
 \label{iteration}
\end{figure}

\textbf{Impact of iteration numbers.}
As shown in Tab.~\ref{iter}, we also conduct experiments with different iteration numbers during the refinement process. 
Ultimately, we notice that the best performance is achieved when the iteration number is set to 2.
It can be observed that during the increase of the iterative times, the finer-grained predictions $P^{ref}$ become increasingly accurate when $i=2$.
However, more iteration times instead result in both high computational costs and gradually degraded performance.
In Fig.~\ref{iteration}, we give examples of predictions $P^{ref}$ refined in the refinement process with iterative feedback. 



\textbf{Impact of different input sizes.}
As different models utilize different input resolutions during COD experiments, we here conduct ablation studies on the input resolution size of model inputs to investigate the effects.
For a fair comparison, we compare our proposed method with the state-of-the-arts with similar input resolutions.
Models settings and corresponding results are given in Tab.~\ref{diffrent size}.
During experiments, we retrain the proposed PCNet on our PlantCamo dataset using input resolution sizes with $384\times384$ and $352\times352$, which are the most commonly used settings in existing COD models.
As shown in Tab.~\ref{diffrent size}, our PCNet achieves state-of-the-art performance under equivalent input resolution settings.
To be specific, when our PCNet is used with a $384\times384$ input, it achieves optimal results, and with a $352\times352$ input, it achieves sub-optimal results. Both of them outperform existing state-of-the-arts.

\begin{table*}[t]
\setlength\tabcolsep{3pt}
\renewcommand{\arraystretch}{1.0}
\footnotesize
\centering
\caption{
Ablation study on input resolution on the PlantCamo Dataset.
The top 3 results are shown in \textbf{\color{red}red}, \textbf{\color{blue}blue}, and \textbf{\color[rgb]{0,0.7,0}green}.}
\label{diffrent size}
\begin{tabular}{l|c|ccccccccc}
\toprule
Method&Resolution&$S_\alpha \uparrow$  & $F^w_\beta \uparrow$ & $M \downarrow$  & $E^{ad}_\varphi \uparrow$ & $E^{m}_\varphi \uparrow$  & $E^{max}_\varphi \uparrow$ & $F^{ad}_\beta \uparrow$& $F^{m}_\beta \uparrow$& $F^{max}_\beta \uparrow$\\
\midrule
\textbf{SINet}~\cite{COD10K}&352$\times$352	&0.766	&0.583	&0.066	&0.842	&0.826	&0.874	&0.664	&0.677	&0.702 \\
\textbf{SINet-V2}~\cite{cod10kpami}&352$\times$352 &0.801 	&0.678 	&0.050 	&0.873 	&0.873 	&0.885 	&0.719 	&0.728 	&0.748 \\
\textbf{ZoomNet}~\cite{ZoomNet}&384$\times$384 &0.798 	&0.680 	&0.049 	&0.874 	&0.858 	&0.875 	&0.725 	&0.731 	&0.744 \\
\textbf{FSPNet}~\cite{FSPNet}&384$\times$384 &0.816	&0.703 	&0.042 	&0.876 	&0.872 	&0.891 	&0.721 	&0.743 	&0.758 \\
\textbf{PFNet}~\cite{PFNet} &416$\times$416	&0.787 	&0.660 	&0.054 	&0.868 	&0.863 	&0.880 	&0.701 	&0.715 	&0.728 \\ 
\textbf{DaCOD}~\cite{dacod} &448$\times$448 &0.804 	&0.693 	&0.046 	&0.889 	&0.879 	&0.895 	&0.729 	&0.745	&0.757 \\
\textbf{UGTR}~\cite{UGTR}&473$\times$473	&0.804	&0.668	&0.050	&0.862	&0.837	&0.883	&0.716	&0.725	&0.747 \\
\textbf{BGNet}~\cite{BGNet} &416$\times$416	&0.786 	&0.552 	&0.076 	&0.873 	&0.846 	&0.889 	&0.723 	&0.716 	&0.757  \\
\textbf{PreyNet}~\cite{PreyNet} &448$\times$448	&0.791 	&0.686 	&0.048 	&0.883 	&0.864 	&0.882 	&0.744 	&0.740 	&0.747  \\
\textbf{SegMaR}~\cite{SegMaR} &352$\times$352	&0.791 	&0.657 	&0.055 	&0.855 	&0.857 	&0.873 	&0.703 	&0.718 	&0.739 \\
\textbf{FAPNet}~\cite{FAPNet} &352$\times$352	&0.801	&0.684	&0.051	&0.879	&0.875	&0.888	&0.729	&0.736	&0.753\\
\textbf{BSANet}~\cite{BSANet}&384$\times$384	&0.782 	&0.663 	&0.053 	&0.875 	&0.851 	&0.878 	&0.718 	&0.720 	&0.729  \\
\textbf{TPRNet}~\cite{TPRNet} &352$\times$352	&0.816&0.680&0.051&0.872&0.883&\textbf{\color{blue}0.910}&0.731&0.750&0.781 \\
\textbf{HitNet}~\cite{HitNet} &384$\times$384 &\textbf{\color[rgb]{0,0.7,0}0.818} 	&\textbf{\color[rgb]{0,0.7,0}0.736} 	&\textbf{\color[rgb]{0,0.7,0}0.041} 	&\textbf{\color[rgb]{0,0.7,0}0.901} 	&\textbf{\color[rgb]{0,0.7,0}0.899} 	&\textbf{\color[rgb]{0,0.7,0}0.902} 	&\textbf{\color[rgb]{0,0.7,0}0.779} 	&\textbf{\color[rgb]{0,0.7,0}0.777} 	&\textbf{\color[rgb]{0,0.7,0}0.784} \\
\midrule
\textbf{PCNet (Ours)}&352$\times$352 &\textbf{\color{blue}0.833} 	&\textbf{\color{blue}0.750} 	&\textbf{\color{blue}0.039} 	&\textbf{\color{blue}0.907} 	&\textbf{\color{blue}0.904} 	&\textbf{\color{blue}0.910} 	&\textbf{\color{blue}0.790} 	&\textbf{\color{blue}0.794} 	&\textbf{\color{blue}0.802} \\
\textbf{PCNet (Ours)}&384$\times$384 &\textbf{\color{red}0.841}	&\textbf{\color{red}0.762}	&\textbf{\color{red}0.037}	&\textbf{\color{red}0.916}	&\textbf{\color{red}0.905}	&\textbf{\color{red}0.915}	&\textbf{\color{red}0.802}	&\textbf{\color{red}0.806}	&\textbf{\color{red}0.813} \\
\bottomrule
\end{tabular}
\end{table*}

\section{Conclusions}\label{con}
In this paper, we present a new perspective on COD by introducing plant camouflage detection and providing the initial effort.
Firstly, we collect the first dataset for plant camouflage, named PlantCamo, which contains 1,250 images with 58 kinds of camouflaged plants.
Based on the PlantCamo dataset, we conduct benchmark studies to understand plant camouflage detection.
Inspired by the plant-specific camouflage characteristics, we further propose a novel PCNet for plant camouflage detection.
PCNet achieves surpassing performance on PlantCamo due to its effective multi-scale feature enhancement and refinement.
Finally, we would like to discuss the applications and future directions.

\textbf{Potential Applications.} PCD has extensive potential applications, especially in biology and agriculture. In biology, it aids in conserving endangered species, identifying new ones, and enhancing our knowledge of herbivore visual systems and plant evolution. Also, AI-driven models facilitate safer and more efficient exploration in challenging environments.
In agriculture, PCD 
can be used for targeted weed control, minimizing environmental impact, and optimizing crop maturity for enhanced yield and quality. These examples underscore PCD's versatility and adaptability in various domains.

\textbf{Limitations and Future Directions.} As we advance in PCD research, it's crucial to recognize existing limitations and explore future avenues. The PlantCamo dataset's limited size and diversity are primary bottlenecks; expanding it with more challenging samples is essential for understanding PCD complexities and enabling advanced algorithms.
In the future, Integrating multiple sensor modalities, 
and importing multi-view approaches can offer valuable insights into PCD by providing unique perspectives and additional clues about plant camouflage.




\bibliographystyle{plain}
\bibliography{neurips_data_2024}

\begin{thebibliography}{10}

\bibitem{F-measure-mean}
Radhakrishna Achanta, Sheila Hemami, Francisco Estrada, and Sabine Susstrunk.
\newblock Frequency-tuned salient region detection.
\newblock In {\em CVPR}, 2009.

\bibitem{C2FNet-V2}
Geng Chen, Si-Jie Liu, Yu-Jia Sun, Ge-Peng Ji, Ya-Feng Wu, and Tao Zhou.
\newblock Camouflaged object detection via context-aware cross-level fusion.
\newblock {\em IEEE TCSVT}, 32(10):6981--6993, 2022.

\bibitem{ASPP}
Liang{-}Chieh Chen, George Papandreou, Iasonas Kokkinos, Kevin Murphy, and Alan~L. Yuille.
\newblock Deeplab: Semantic image segmentation with deep convolutional nets, atrous convolution, and fully connected crfs.
\newblock {\em {IEEE} TPAMI}, 40(4):834--848, 2018.

\bibitem{boundary1}
Tianyou Chen, Jin Xiao, Xiaoguang Hu, Guofeng Zhang, and Shaojie Wang.
\newblock Boundary-guided network for camouflaged object detection.
\newblock {\em KBS}, 248:108901, 2022.

\bibitem{camodiffusion}
Zhongxi Chen, Ke~Sun, and Xianming Lin.
\newblock Camodiffusion: Camouflaged object detection via conditional diffusion models.
\newblock In {\em AAAI}, 2024.

\bibitem{FPNet}
Runmin Cong, Mengyao Sun, Sanyi Zhang, Xiaofei Zhou, Wei Zhang, and Yao Zhao.
\newblock Frequency perception network for camouflaged object detection.
\newblock In {\em ACM MM}, 2023.

\bibitem{UQFormer}
Bo~Dong, Jialun Pei, Rongrong Gao, Tian{-}Zhu Xiang, Shuo Wang, and Huan Xiong.
\newblock A unified query-based paradigm for camouflaged instance segmentation.
\newblock In {\em ACM MM}, 2023.

\bibitem{ViT}
Alexey Dosovitskiy, Lucas Beyer, Alexander Kolesnikov, Dirk Weissenborn, Xiaohua Zhai, Thomas Unterthiner, Mostafa Dehghani, Matthias Minderer, Georg Heigold, Sylvain Gelly, Jakob Uszkoreit, and Neil Houlsby.
\newblock An image is worth 16x16 words: Transformers for image recognition at scale.
\newblock In {\em {ICLR}}, 2021.

\bibitem{dyer2021plant}
Adrian~G Dyer and Jair~E Garcia.
\newblock Plant camouflage: fade to grey.
\newblock {\em Curr. Biol.}, 31(2):R78--R80, 2021.

\bibitem{S-measure}
Deng-Ping Fan, Ming-Ming Cheng, Yun Liu, Tao Li, and Ali Borji.
\newblock Structure-measure: A new way to evaluate foreground maps.
\newblock In {\em ICCV}, 2017.

\bibitem{cod10kpami}
Deng{-}Ping Fan, Ge{-}Peng Ji, Ming{-}Ming Cheng, and Ling Shao.
\newblock Concealed object detection.
\newblock {\em {IEEE} TPAMI}, 44(10):6024--6042, 2022.

\bibitem{E-measure}
Deng-Ping Fan, Ge-Peng Ji, Xuebin Qin, and Ming-Ming Cheng.
\newblock Cognitive vision inspired object segmentation metric and loss function.
\newblock {\em SSI}, 6(6), 2021.

\bibitem{COD10K}
Deng-Ping Fan, Ge-Peng Ji, Guolei Sun, Ming-Ming Cheng, Jianbing Shen, and Ling Shao.
\newblock Camouflaged object detection.
\newblock In {\em CVPR}, 2020.

\bibitem{advances}
Deng{-}Ping Fan, Ge{-}Peng Ji, Peng Xu, Ming{-}Ming Cheng, Christos Sakaridis, and Luc~Van Gool.
\newblock Advances in deep concealed scene understanding.
\newblock {\em CoRR}, abs/2304.11234, 2023.

\bibitem{Res2Net}
Shang-Hua Gao, Ming-Ming Cheng, Kai Zhao, Xin-Yu Zhang, Ming-Hsuan Yang, and Philip Torr.
\newblock Res2net: A new multi-scale backbone architecture.
\newblock {\em IEEE TPAMI}, page 652–662, 2021.

\bibitem{FEDER}
Chunming He, Kai Li, Yachao Zhang, Longxiang Tang, Yulun Zhang, Zhenhua Guo, and Xiu Li.
\newblock Camouflaged object detection with feature decomposition and edge reconstruction.
\newblock In {\em CVPR}, 2023.

\bibitem{iclr24}
Chunming He, Kai Li, Yachao Zhang, Yulun Zhang, Chenyu You, Zhenhua Guo, Xiu Li, Martin Danelljan, and Fisher Yu.
\newblock Strategic preys make acute predators: Enhancing camouflaged object detectors by generating camouflaged objects.
\newblock In {\em {ICLR}}. OpenReview.net, 2024.

\bibitem{ResNet}
Kaiming He, Xiangyu Zhang, Shaoqing Ren, and Jian Sun.
\newblock Deep residual learning for image recognition.
\newblock In {\em CVPR}, 2016.

\bibitem{CRNet}
Ruozhen He, Qihua Dong, Jiaying Lin, and Rynson W.~H. Lau.
\newblock Weakly-supervised camouflaged object detection with scribble annotations.
\newblock In {\em AAAI}, 2023.

\bibitem{HitNet}
Xiaobin Hu, Shuo Wang, Xuebin Qin, Hang Dai, Wenqi Ren, Donghao Luo, Ying Tai, and Ling Shao.
\newblock High-resolution iterative feedback network for camouflaged object detection.
\newblock In {\em AAAI}, 2023.

\bibitem{FSPNet}
Zhou Huang, Hang Dai, Tian-Zhu Xiang, Shuo Wang, Huai-Xin Chen, Jie Qin, and Huan Xiong.
\newblock Feature shrinkage pyramid for camouflaged object detection with transformers.
\newblock In {\em CVPR}, 2023.

\bibitem{DGNet}
Ge-Peng Ji, Deng-Ping Fan, Yu-Cheng Chou, Dengxin Dai, Alexander Liniger, and Luc Van~Gool.
\newblock Deep gradient learning for efficient camouflaged object detection.
\newblock {\em MIR}, 20(1):92--108, 2023.

\bibitem{ERRNet}
Ge{-}Peng Ji, Lei Zhu, Mingchen Zhuge, and Keren Fu.
\newblock Fast camouflaged object detection via edge-based reversible re-calibration network.
\newblock {\em PR}, 2022.

\bibitem{SegMaR}
Qi~Jia, Shuilian Yao, Yu~Liu, Xin Fan, Risheng Liu, and Zhongxuan Luo.
\newblock Segment, magnify and reiterate: Detecting camouflaged objects the hard way.
\newblock In {\em CVPR}, 2022.

\bibitem{lamdouar2023making}
Hala Lamdouar, Weidi Xie, and Andrew Zisserman.
\newblock The making and breaking of camouflage.
\newblock In {\em ICCV}, 2023.

\bibitem{CAMO}
Trung-Nghia Le, Tam~V. Nguyen, Zhongliang Nie, Minh-Triet Tran, and Akihiro Sugimoto.
\newblock Anabranch network for camouflaged object segmentation.
\newblock {\em CVIU}, 2019.

\bibitem{JSCOD}
Aixuan Li, Jing Zhang, Yunqiu Lv, Bowen Liu, Tong Zhang, and Yuchao Dai.
\newblock Uncertainty-aware joint salient object and camouflaged object detection.
\newblock In {\em CVPR}, 2021.

\bibitem{findnet}
Peng Li, Xuefeng Yan, Hongwei Zhu, Mingqiang Wei, Xiao{-}Ping Zhang, and Jing Qin.
\newblock Findnet: Can you find me? boundary-and-texture enhancement network for camouflaged object detection.
\newblock {\em {IEEE} TIP}, 31:6396--6411, 2022.

\bibitem{penet}
Xiaofei Li, Jiaxin Yang, Shuohao Li, Jun Lei, Jun Zhang, and Dong Chen.
\newblock Locate, refine and restore: {A} progressive enhancement network for camouflaged object detection.
\newblock In {\em {IJCAI}}, 2023.

\bibitem{OCENet}
Jiawei Liu, Jing Zhang, and Nick Barnes.
\newblock Modeling aleatoric uncertainty for camouflaged object detection.
\newblock In {\em WACV}, 2022.

\bibitem{EVP}
Weihuang Liu, Xi~Shen, Chi{-}Man Pun, and Xiaodong Cun.
\newblock Explicit visual prompting for low-level structure segmentations.
\newblock In {\em CVPR}, 2023.

\bibitem{Swin}
Ze~Liu, Yutong Lin, Yue Cao, Han Hu, Yixuan Wei, Zheng Zhang, Stephen Lin, and Baining Guo.
\newblock Swin transformer: Hierarchical vision transformer using shifted windows.
\newblock In {\em ICCV}, 2022.

\bibitem{DTINet}
Zhengyi Liu, Zhili Zhang, Yacheng Tan, and Wei Wu.
\newblock Boosting camouflaged object detection with dual-task interactive transformer.
\newblock In {\em ICPR}, 2022.

\bibitem{vscode}
Ziyang Luo, Nian Liu, Wangbo Zhao, Xuguang Yang, Dingwen Zhang, Deng-Ping Fan, Fahad Khan, and Junwei Han.
\newblock Vscode: General visual salient and camouflaged object detection with 2d prompt learning.
\newblock In {\em CVPR}, 2024.

\bibitem{NC4K}
Yunqiu Lv, Jing Zhang, Yuchao Dai, Aixuan Li, Bowen Liu, Nick Barnes, and Deng-Ping Fan.
\newblock Simultaneously localize, segment and rank the camouflaged objects.
\newblock In {\em CVPR}, 2021.

\bibitem{F-measure}
Ran Margolin, Lihi Zelnik-Manor, and Ayellet Tal.
\newblock How to evaluate foreground maps.
\newblock In {\em CVPR}, 2014.

\bibitem{PFNet}
Haiyang Mei, Ge-Peng Ji, Ziqi Wei, Xin Yang, Xiaopeng Wei, and Deng-Ping Fan.
\newblock Camouflaged object segmentation with distraction mining.
\newblock In {\em CVPR}, 2021.

\bibitem{niu2017divergence}
Yang Niu, Zhe Chen, Martin Stevens, and Hang Sun.
\newblock Divergence in cryptic leaf colour provides local camouflage in an alpine plant.
\newblock {\em Proc. Royal Soc. B}, 284(1864):20171654, 2017.

\bibitem{commercial}
Yang Niu, Martin Stevens, and Hang Sun.
\newblock Commercial harvesting has driven the evolution of camouflage in an alpine plant.
\newblock {\em Curr. Biol.}, 31(2):446--449, 2021.

\bibitem{Niu2014AlpineSP}
Yang Niu and Hang Sun.
\newblock Alpine scree plants benefit from cryptic coloration with limited cost.
\newblock {\em Plant Signal. Behav.}, 9:63, 2014.

\bibitem{plant}
Yang Niu, Hang Sun, and Martin Stevens.
\newblock Plant camouflage: ecology, evolution, and implications.
\newblock {\em TREE}, 33(8):608--618, 2018.

\bibitem{ZoomNet}
Youwei Pang, Xiaoqi Zhao, Tian-Zhu Xiang, Lihe Zhang, and Huchuan Lu.
\newblock Zoom in and out: A mixed-scale triplet network for camouflaged object detection.
\newblock In {\em CVPR}, 2022.

\bibitem{chameleon}
Przemys{\l}aw Skurowski, Hassan Abdulameer, J~B{\l}aszczyk, Tomasz Depta, Adam Kornacki, and P~Kozie{\l}.
\newblock Animal camouflage analysis: Chameleon database.
\newblock {\em Unpublished manuscript}, 2(6):7, 2018.

\bibitem{C2FNet}
Yujia Sun, Geng Chen, Tao Zhou, Yi~Zhang, and Nian Liu.
\newblock Context-aware cross-level fusion network for camouflaged object detection.
\newblock In {\em IJCAI}, 2021.

\bibitem{BGNet}
Yujia Sun, Shuo Wang, Chenglizhao Chen, and Tian-Zhu Xiang.
\newblock Boundary-guided camouflaged object detection.
\newblock In {\em IJCAI}, 2022.

\bibitem{Eff}
Mingxing Tan and Quoc~V. Le.
\newblock Efficientnet: Rethinking model scaling for convolutional neural networks.
\newblock In {\em {ICML}}, 2019.

\bibitem{Color_of_Animal_plants}
Alfred~Russel Wallace.
\newblock The colors of animals and plants.
\newblock {\em Am. Nat.}, 11(11):641--662, 1877.

\bibitem{dacod}
Qingwei Wang, Jinyu Yang, Xiaosheng Yu, Fangyi Wang, Peng Chen, and Feng Zheng.
\newblock Depth-aided camouflaged object detection.
\newblock In {\em {ACM} MM}, 2023.

\bibitem{pvtv2}
Wenhai Wang, Enze Xie, Xiang Li, Deng{-}Ping Fan, Kaitao Song, Ding Liang, Tong Lu, Ping Luo, and Ling Shao.
\newblock {PVT} v2: Improved baselines with pyramid vision transformer.
\newblock {\em CVM}, 8(3):415--424, 2022.

\bibitem{MENet}
Yi~Wang, Ruili Wang, Xin Fan, Tianzhu Wang, and Xiangjian He.
\newblock Pixels, regions, and objects: Multiple enhancement for salient object detection.
\newblock In {\em CVPR}, 2023.

\bibitem{popnet}
Zongwei Wu, Danda~Pani Paudel, Deng-Ping Fan, Jingjing Wang, Shuo Wang, C{\'e}dric Demonceaux, Radu Timofte, and Luc Van~Gool.
\newblock Source-free depth for object pop-out.
\newblock In {\em ICCV}, 2023.

\bibitem{FRINet}
Chenxi Xie, Changqun Xia, Tianshu Yu, and Jia Li.
\newblock Frequency representation integration for camouflaged object detection.
\newblock In {\em ACM MM}, 2023.

\bibitem{MiT}
Enze Xie, Wenhai Wang, Zhiding Yu, Anima Anandkumar, Jos{\'{e}}~M. {\'{A}}lvarez, and Ping Luo.
\newblock Segformer: Simple and efficient design for semantic segmentation with transformers.
\newblock In {\em NeurIPS}, 2021.

\bibitem{UGTR}
Fan Yang, Qiang Zhai, Xin Li, Rui Huang, Ao~Luo, Hong Cheng, and Deng-Ping Fan.
\newblock Uncertainty-guided transformer reasoning for camouflaged object detection.
\newblock In {\em ICCV}, 2022.

\bibitem{camoformer}
Bowen Yin, Xuying Zhang, Qibin Hou, Bo{-}Yuan Sun, Deng{-}Ping Fan, and Luc~Van Gool.
\newblock Camoformer: Masked separable attention for camouflaged object detection.
\newblock {\em CoRR}, abs/2212.06570, 2022.

\bibitem{highreso}
Yi~Zeng, Pingping Zhang, Zhe~L. Lin, Jianming Zhang, and Huchuan Lu.
\newblock Towards high-resolution salient object detection.
\newblock In {\em ICCV}, 2019.

\bibitem{MGL}
Qiang Zhai, Xin Li, Fan Yang, Chenglizhao Chen, Hong Cheng, and Deng-Ping Fan.
\newblock Mutual graph learning for camouflaged object detection.
\newblock In {\em CVPR}, 2021.

\bibitem{depthcod}
Jing Zhang, Yunqiu Lv, Mochu Xiang, Aixuan Li, Yuchao Dai, and Yiran Zhong.
\newblock Depth-guided camouflaged object detection.
\newblock {\em CoRR}, abs/2106.13217, 2021.

\bibitem{PreyNet}
Miao Zhang, Shuang Xu, Yongri Piao, Dongxiang Shi, Shusen Lin, and Huchuan Lu.
\newblock Preynet: Preying on camouflaged objects.
\newblock In {\em ACM MM}, 2022.

\bibitem{TPRNet}
Qiao Zhang, Yanliang Ge, Cong Zhang, and Hongbo Bi.
\newblock Tprnet: camouflaged object detection via transformer-induced progressive refinement network.
\newblock {\em VC}, pages 1--15, 2022.

\bibitem{MFFN}
Dehua Zheng, Xiaochen Zheng, Laurence~T. Yang, Yuan Gao, Chenlu Zhu, and Yiheng Ruan.
\newblock {MFFN:} multi-view feature fusion network for camouflaged object detection.
\newblock In {\em WACV}, 2023.

\bibitem{FDNet}
Yijie Zhong, Bo~Li, Lv~Tang, Senyun Kuang, Shuang Wu, and Shouhong Ding.
\newblock Detecting camouflaged object in frequency domain.
\newblock In {\em CVPR}, 2022.

\bibitem{FAPNet}
Tao Zhou, Yi~Zhou, Chen Gong, Jian Yang, and Yu~Zhang.
\newblock Feature aggregation and propagation network for camouflaged object detection.
\newblock {\em IEEE TIP}, 31:7036--7047, 2022.

\bibitem{BSANet}
Hongwei Zhu, Peng Li, Haoran Xie, Xuefeng Yan, Dong Liang, Dapeng Chen, Mingqiang Wei, and Jing Qin.
\newblock I can find you! boundary-guided separated attention network for camouflaged object detection.
\newblock In {\em AAAI}, 2022.

\bibitem{TINet}
Jinchao Zhu, Xiaoyu Zhang, Shuo Zhang, and Junnan Liu.
\newblock Inferring camouflaged objects by texture-aware interactive guidance network.
\newblock In {\em {AAAI}}, 2021.

\end{thebibliography}

\end{document}